%% file: main.tex
\documentclass[11pt]{article}
\usepackage[preprint]{acl}
\usepackage{times}
\usepackage{latexsym}
\usepackage[T1]{fontenc}
\usepackage[utf8]{inputenc}
\usepackage{microtype}
\usepackage{graphicx}
\usepackage{booktabs}
\usepackage{array}
\newcolumntype{L}[1]{>{\raggedright\arraybackslash}p{#1}}
\usepackage{multirow}
\usepackage{amsmath}
\usepackage{amssymb}
\usepackage{xcolor}
\usepackage{colortbl}  
\usepackage{textcomp}  

\newcommand{\U}{U}
\newcommand{\D}{D}

\title{Reading Right, Acting Wrong:\\
Diagnosing Decision Failures of System One Models in Agent Loops\thanks{Project website: \url{https://yubol-bobo.github.io/System_one_model_evaluation}}}

\author{%
Yubo Li \quad Ramayya Krishnan \quad Rema Padman\\[3pt]
\mdseries Carnegie Mellon University\\
\mdseries\texttt{\{yubol, rk2x, rpadman\}@andrew.cmu.edu}\\[3pt]
}

\begin{document}
\maketitle

\input{sections/00_abstract}
\input{sections/01_introduction}
\input{sections/03_setting}
\input{sections/04_benchmark}
\input{sections/05_systems}

\input{sections/06_results}
\input{sections/07_mechanism}
\input{sections/08_controls}
\input{sections/02_related}
\input{sections/09_discussion}
\input{sections/10_closing}

{\renewcommand{\href}[2]{#2}
\bibliography{references}}

\appendix
\input{sections/appendix_related}
\input{sections/appendix_measurement}
\input{sections/appendix_benchmark}
\input{sections/appendix_protocol}

\input{sections/appendix}

\end{document}

%% file: sections/00_abstract.tex
\begin{abstract}
System One models choose an agent's next action by scoring a supplied set of options in a single pass. This
makes them fast and cheap, but when they choose wrongly there is nothing to inspect: did the model misread
what the agent has seen, or read it correctly and still act wrongly? We answer behaviorally. On a frozen
benchmark of agent decision states from simulated environments, customer-service policies, tool use, web
navigation and coding, we pair every next-action question with factual probes about the same state, each
answered independently. Across seventeen fast systems and a reasoning model, the answer is consistent: most
wrong actions follow correct reading, and the more accurate a system, the more of its errors take this
form. These errors are not noise; they gather on the same states across systems of every kind. Where the
probed facts provably determine the action, models that read every fact still fail to combine them, and
their own answers, composed by code, repair the failure. On real states, what is missing is often not a
fact but the next sub-goal.
\end{abstract}

%% file: sections/01_introduction.tex
\section{Introduction}
\label{sec:intro}

\begin{figure*}[!tp]
\centering
\includegraphics[width=\textwidth]{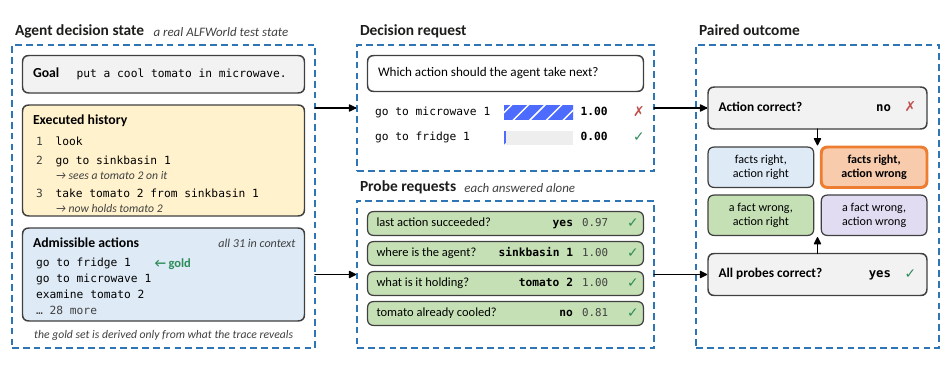}
\caption{Paired measurement on a real ALFWorld state. Left: the goal, the history and all 31 admissible actions; the
gold action follows only from what the trace reveals. Middle: the decision and each factual probe are answered
independently on the same state (JEV's answers: every probe right, action wrong). Right: crossing all facts correct
($\U$) with action correct ($\D$) gives four outcomes; this state is facts right, action wrong.}
\label{fig:overview}
\end{figure*}

An agent acting in an environment repeatedly reads the feedback of its previous actions, works out where it
stands, and chooses what to do next. In most current agents each such step is a call to a large language model
that writes out its reasoning and then names an action \citep{yao2023react}. A growing family of \emph{System One
models} takes a different route \citep{typesafe2026jev,liquid2026d1,clm2026,kev2026,laya2026,clef2026}. Given the
agent's state and a set of candidate answers, they return a probability distribution over the supplied options and
generate no text; their output is consumed directly by code, and their developers emphasize intelligence per
dollar \citep{typesafe2026jev}. The System One models we evaluate answer a complete state in a median of
27--225\,ms, against 317--947\,ms for likelihood readouts of open-weight language models and 2.3\,s for a reasoning
model (\S\ref{sec:latency}). We study only their \emph{choice interface}, in which the model selects among supplied
options.

When such a model picks the wrong action, there is no chain of thought to read. The error has two very different
explanations. The model may have \emph{misread} what the agent already knows, for example by missing that the last
action failed or losing track of what it holds; or it may have read all of it correctly and still \emph{chosen
badly}. The first points to the state representation or the context budget, the second to the decision itself.
Final-action accuracy cannot tell these cases apart, and generated rationales need not reflect the computation
behind an answer \citep{turpin2023unfaithful,lanham2023faithfulness}. Conversely, a correct action does not show
that the model understood the trace.

We therefore ask, behaviorally: \emph{can System One models understand the information an agent has obtained and
act correctly on it, and when they fail, can we tell failures of comprehension from failures of decision?} On a
single agent decision state we ask, each in isolation, the next-action question and factual probes whose answers
the environment determines, such as whether the last action succeeded, what the agent carries, or whether an action
is permitted. Crossing ``all measured facts correct'' with ``action correct'' gives four observable outcomes
(Figure~\ref{fig:overview}). Around this measurement we build a frozen benchmark of 65,228 records, 19,663 of them
paired test states, from simulators with exact ground truth (ALFWorld, BabyAI-Text, TextWorld), customer-service
policies ($\tau^2$-bench), tool use, web navigation and coding traces, with every admissible action in context and
gold labels computed only from what the agent has seen. We evaluate seventeen fast systems, fourteen System One models
of three kinds (hosted APIs, open-weight models, and recipes that read a frozen language model) and three likelihood
readouts of open-weight language models, with a reasoning model, GPT-6.1, as a reference.

The short answer is that these models mostly read right and act wrong. Our main findings are:
\begin{itemize}
\item \textbf{Most wrong actions follow correct reading.} For the six most accurate systems, 73--85\% of decision
errors occur on states where every measured fact is answered correctly, and the share grows with accuracy
(\S\ref{sec:dissociation}).
\item \textbf{The errors are structured.} All 14 competent fast systems fail with every fact right on the same 141
states, where chance predicts none (\S\ref{sec:consistency}).
\item \textbf{When the probes suffice, the failure is composition.} On generated states whose probes determine the
action, models that read every fact and decide perfectly from the facts alone lose 20--45 points when they must do
both from the trace. Their own fact answers recover 15--28 points, and composing the action in code from them
reaches 0.96--1.00 (\S\ref{sec:ladder}).
\item \textbf{On real states, the decision itself is often missing.} Correct answers to the benchmark's probes add
at most 7 points for all but two systems. On ALFWorld, stating the next sub-goal adds 16 points for JEV, any block of
facts at most 5 (\S\ref{sec:ptd}, \S\ref{sec:dprobes}).
\item \textbf{Accuracy hides the rest.} Removing the history collapses accuracy where earlier observations matter,
policy edits are read correctly but acted on in only 22--72\% of cases, and calibration error ranges from 2.6\% to
26.6\% (\S\ref{sec:controls-results}).
\end{itemize}
We also contribute a model-free criterion for when probes can support attributing an error to the decision
(\S\ref{sec:sufficiency}), states that satisfy it by construction, and an evaluation of eighteen systems that
combines the paired diagnostics with cost, latency and calibration. The paired outcomes are behavioral evidence: they
do not identify a model's internal computation, and we do not claim that the probes exhaust every relevant fact.

%% file: sections/03_setting.tex
\section{Measuring Comprehension and Decision on the Same State}
\label{sec:setting}

\subsection{Decision states}
\label{sec:states}

A decision state $s$ consists of a task goal, the executed history of actions and the observations they produced,
any rules that govern the environment or policy, and the set $\mathcal{A}(s)$ of admissible next actions. The model
must choose one element of $\mathcal{A}(s)$ and is never asked to generate an action. Every admissible action is
placed in context with its name and a description, nothing is shortlisted, and no language model proposes
candidates; tool calls with open arguments are enumerated from the environment or database. Gold is a \emph{set}
$G(s)\subseteq\mathcal{A}(s)$ computed only from information available to the agent: in the simulators, the first
actions of all shortest plans on a map built from what the trace entails, or, when the needed objects have not yet
been seen, the actions that reveal unseen locations. A decision is correct, $\D(s)=1$, if the chosen action lies in
$G(s)$.

\subsection{Factual probes}
\label{sec:probes}

On annotated states we also ask closed factual probes whose answers the simulator state, database or data labels
determine: the outcome of the last action, the agent's location and inventory, completed subgoals, whether an action
is permitted, and which observation establishes a fact. Each question, probe or decision, is answered on the state
alone; several may share a request, but no question or answer is added to another question's input and no gold fact
is revealed. ``Paired'' means that the input state is identical, not that we observe stages of the model's
computation.

\subsection{Four observable outcomes}
\label{sec:outcomes}

Let $\U_i=1$ if all applicable probes of state $i$ are answered correctly. $\U_i$ is measured-fact performance, not
complete understanding. Crossing $\U$ with $\D$ gives four outcomes; we write U1D0 for $\U{=}1,\D{=}0$ (facts right,
action wrong), and U1D1, U0D1, U0D0 likewise. Each licenses less than it may seem (Table~\ref{tab:outcomes} in
Appendix~\ref{app:setting}); in particular, U1D0 does not show that the failure lies purely in deciding, and U0D1 does
not show a shortcut. Questions a system cannot answer because of an input limit are counted wrong and reported as
coverage. We report counts pooled over the 19,663 paired test states, which summarize prevalence and are not a ranking
statistic (cell-weighted estimates and cluster bootstrap: Appendix~\ref{app:outcomes}).

\subsection{When can a probe set support the diagnosis?}
\label{sec:sufficiency}

A U1D0 outcome suggests a decision failure only if the probes capture what the decision requires, and this can be
checked without the model. Holding goal, rules and remaining context fixed, let $P(s)$ be the vector of correct probe
answers. For a group of states $S$ sharing a probe vector, the probe set is \emph{sufficient on} $S$ if
\begin{equation}
\textstyle\bigcap_{s\in S} G(s)\neq\emptyset,
\label{eq:sufficiency}
\end{equation}
that is, if some action is correct in every state the agent could be in given those answers. If two states share
the correct probe answers but have disjoint gold sets, answering every probe correctly cannot determine the action,
and a wrong action after correct probes cannot be attributed to deciding alone. Pairwise intersections are not
enough; the check must consider the whole group. In finite environments it is exhaustive; in natural data, absence
of a counterexample does not prove sufficiency. Passing it means the probes \emph{can} support a diagnosis, not that
the model uses those facts.

\subsection{Controlled variations}
\label{sec:controls}

To see what drives the outcomes, the benchmark also contains a trace ablation that removes the executed history,
observation minimal pairs that edit only the last observation relevantly or irrelevantly, stated rule changes that
make prior knowledge wrong, and difficulty ladders over trace length and the distance from the decisive observation
(\S\ref{sec:controls-results}); encoding controls are not analyzed here (Appendix~\ref{app:controls}).

%% file: sections/04_benchmark.tex
\section{Benchmark}
\label{sec:benchmark}

The frozen benchmark has 65,228 records, 59,234 test and 5,994 dev (Table~\ref{tab:composition} in
Appendix~\ref{app:benchmark}). Most paired states come from three simulators with exact ground truth: ALFWorld,
BabyAI-Text and TextWorld cooking games
\citep{shridhar2021alfworld,paglieri2025balrog,chevalierboisvert2019babyai,carta2023glam,cote2018textworld,tales2025}.
Trajectories are rolled out by each environment's expert, by a noisy variant and by a variant that issues invalid
commands, so traces contain failures and detours; no language model takes part. Gold sets were checked against
breadth-first search in the real environments. From $\tau^2$-bench airline and retail \citep{barres2025tau2bench} we
take policy-constrained states with tool calls enumerated from the database and probes on record fields and
permissions. Tool selection, web navigation and AgentRM trees
\citep{toucan2025,shi2025toolret,patil2025bfcl,ross2025when2call,deng2023mind2web,xia2025agentrm} carry probes on
fields stated in the trace (the \emph{visible-field} probes), and 360 coding records built from real repository
histories ask which edit caused a regression and which limited next action follows. Ladders vary trace length,
distance from the decisive observation and novelty, and confirmation cohorts generated later from new seeds re-test
the findings on states no earlier run had seen. A further 32,700 records of text-understanding primitives are
part of the frozen population but not analyzed here.

Records are split into dev and test by source (game, user, repository), and every request byte, option order and
label is frozen in a hash-sealed manifest before analysis. An independent content audit flagged 98 test records,
which we keep and treat in a sensitivity analysis (\S\ref{sec:labels}). Because fixed public items can be overfit,
the follow-up studies also use freshly generated states, including states whose probes are sufficient by
construction (\S\ref{sec:ladder}).

%% file: sections/05_systems.tex
\section{Systems and Protocol}
\label{sec:systems}

\input{tables/systems}

\paragraph{Systems.} Table~\ref{tab:systems} lists the seventeen fast systems and the reasoning reference, grouped by
type. The fourteen System One models return a native distribution over the supplied options in one pass: two hosted
APIs, JEV and OpenAI Decisions; ten released open-weight models, served locally from pinned revisions with their
authors' code and settings; and two recipes that read a frozen open-weight language model through the same
interface. Gemma-4-31B, Qwen3.8-27B and Qwen3.5-9B are read out by complete-option likelihood with thinking disabled,
which gives a distribution of the same form. GPT-6.1 Sol (medium reasoning) answers each question in a separate call
with the selected option and a verbalized distribution; it is a reference with deliberation, not a competitor.

\paragraph{Identical inputs.} Every system receives the same frozen requests, and we never truncate a state, drop
options or reselect records. A question beyond a system's documented input limit is recorded as unsupported, counted
wrong and reported as coverage; this matters mainly for Laya, Quyet-1.0-Large (which refuses more than ten options,
including every ALFWorld decision), Perplexity Decider-27B, and the servers of Vega-27B and SimpleJev-27B on the
longest states (Appendix~\ref{app:systems}). A system that answers fewer than half of the questions of an analysis is
left out of it and named (the \emph{coverage rule}). Readouts differ by design, so comparisons are between deployed
choice interfaces and do not isolate architecture, size or training data. All systems receive all 65,228 records.

\paragraph{Latency.}\label{sec:latency} Accuracy runs batch requests, so we time a separate \emph{latency panel} of
163 test items, one per difficulty cell, answered strictly one item at a time after five warm-up items; we report the
first pass on complete answers (Table~\ref{tab:systems}). Systems marked there are estimated from their sequential
accuracy runs; GPT-6.1 is timed through its synchronous API with an item's questions sent in parallel
(Appendix~\ref{app:latency}).

\begin{figure*}[!tp]
\centering
\includegraphics[width=\textwidth]{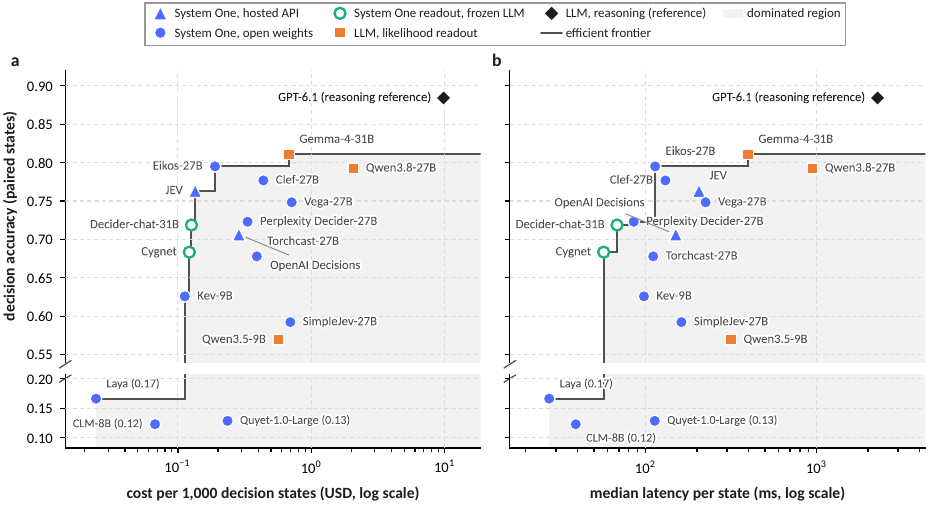}
\caption{Paired decision accuracy against (a) cost per 1,000 decision states and (b) median latency. The step line is
the efficient frontier. Hosted costs are measured billing; self-hosted costs assume \$2.50 per H100-hour on the
latency panel, an upper bound for batched serving (\S\ref{sec:latency}). Markers give the system category; Laya,
Quyet-1.0-Large and CLM-8B lie below the axis break.}
\label{fig:pareto}
\end{figure*}

\paragraph{Cost.}\label{sec:efficiency} JEV was billed \$8.79 for all 65,228 records (\$0.13 per 1,000 decision states)
and OpenAI Decisions \$18.69 (\$0.29). For self-hosted systems we estimate cost from the panel's time per state at a
stated GPU price. The efficient frontier (Figure~\ref{fig:pareto}) runs from Laya and Kev-9B through Cygnet,
Decider-chat-31B and JEV to Eikos-27B, which reaches 0.795 at about 1.4 times JEV's cost, and Gemma-4-31B (0.811 at
about five times). GPT-6.1 reaches 0.884 for \$642 under batch pricing, about 73 times JEV per state. Hosted models
can change between versions: a rerun of JEV's decisions five days after the formal run matched on 96.3\% of 2,783
states (Appendix~\ref{app:systems}).

%% file: tables/systems.tex
\begin{table*}[!tp]
\centering
\footnotesize
\setlength{\tabcolsep}{4pt}
\renewcommand{\arraystretch}{1.12}
\begin{tabular}{@{}L{3.4cm}L{2.95cm}L{3.45cm}L{2.75cm}rr@{}}
\toprule
 & & & & \multicolumn{2}{c}{Latency (ms)} \\
\cmidrule(l){5-6}
System & Model & Readout & Input limit & p50 & p95 \\
\midrule
\multicolumn{6}{@{}l}{\emph{System One models, hosted API}} \\
JEV 1.13 \citep{typesafe2026jev} & hosted & native distribution & 32k & 205.7 & 290.5 \\
OpenAI Decisions \citep{openai2026decisions} & hosted (public beta) & native distribution (two decimals) & none documented & 150.6$^\ast$ & 197.6$^\ast$ \\
\midrule
\multicolumn{6}{@{}l}{\emph{System One models, open weights}} \\
Clef-27B \citep{clef2026} & Qwen3.8-27B, post-trained & native schema head & 32k$^\dagger$ & 130.9 & 1,990.5 \\
Kev-9B \citep{kev2026} & Qwen3.5-9B-Base & pointer head & 64k state & 98.1 & 562.0 \\
CLM-8B \citep{clm2026} & Qwen3-8B embeddings & two-tower head & 32k per text$^\dagger$ & 39.2 & 475.9 \\
Laya \citep{laya2026} & encoder & native distribution & 512 & 27.4 & 69.8 \\
Eikos-27B \citep{eikos2026} & Qwen3.8-27B fine-tune, FP8 & letter readout & 32k$^\dagger$ & 114.0$^\ast$ & 1,117.8$^\ast$ \\
Perplexity Decider-27B \citep{pplxdecider2026} & Qwen3.8-27B fine-tune & readout head & 8k per question & 85.4$^\ast$ & 1,294.3$^\ast$ \\
SimpleJev-27B \citep{simplejev2026} & Qwen3.8-27B + LoRA & pointer head & 32k$^\dagger$; GPU memory$^\S$ & 162.4$^\ast$ & 3,798.9$^\ast$ \\
Vega-27B \citep{vega2026} & Qwen3.8-27B + adapter & Decision 2.0 runtime & 32k; GPU memory$^\S$ & 224.9$^\ast$ & 3,756.8$^\ast$ \\
Torchcast-27B \citep{torchcast2026} & StartLux-Decision-27B fine-tune & letter readout & 32k$^\dagger$ & 111.1$^\ast$ & 2,164.5$^\ast$ \\
Quyet-1.0-Large \citep{quyet2026} & Gemma-4-31B-it + LoRA & letter readout & state cut at 6k; $\le$10 options$^\ddagger$ & 113.4$^\ast$ & 1,297.3$^\ast$ \\
\midrule
\multicolumn{6}{@{}l}{\emph{System One readout of a frozen LLM}} \\
Cygnet \citep{cygnet2026} & Gemma-4-12B-it & grouped letter probabilities & 16k & 57.0 & 698.0 \\
Decider-chat-31B \citep{deciderchat2026} & Gemma-4-31B-it & letter readout & 8k; state cut at 4k$^\ddagger$ & 68.3$^\ast$ & 550.6$^\ast$ \\
\midrule
\multicolumn{6}{@{}l}{\emph{LLM, likelihood readout (thinking off)}} \\
Gemma-4-31B \citep{gemma4} & Gemma-4-31B-it & option likelihood, thinking off & 32k & 397.9 & 3,476.5 \\
Qwen3.8-27B \citep{qwen38} & Qwen3.8-27B & option likelihood, thinking off & 32k & 947.0 & 12,598.8 \\
Qwen3.5-9B \citep{qwen35} & Qwen3.5-9B & option likelihood, thinking off & 32k & 317.0 & 2,502.7 \\
\midrule
\multicolumn{6}{@{}l}{\emph{LLM, reasoning (reference)} (evaluated per question)} \\
GPT-6.1 Sol \citep{openai2026gpt61} & hosted reasoning model & answer + verbalized distribution & -- & 2,276.0$^\P$ & 5,756.8$^\P$ \\
\bottomrule
\end{tabular}
\caption{Systems by type, with readouts, input limits and median and 95th-percentile latency on the latency panel
(\S\ref{sec:latency}; local systems on one H100). $^\ast$Estimated from sequential accuracy runs. $^\dagger$Our
explicit budget (release defaults: 16k for Clef-27B and Eikos-27B, 8k for SimpleJev-27B, 256k for Torchcast-27B; CLM-8B's serving example uses 2k). $^\ddagger$Applied by the author's runtime: Decider-chat-31B cuts states
beyond 4,096 tokens; Quyet-1.0-Large cuts states to at most 6,000 tokens and refuses more than ten options.
$^\S$The server fails on the longest states on one 80\,GB H100. $^\P$Synchronous API, one question per request, an item's questions sent in parallel.}
\label{tab:systems}
\end{table*}

%% file: sections/06_results.tex
\section{Reading Correctly Is Not Enough}
\label{sec:results}

\input{tables/paired}

\begin{figure*}[!tp]
\centering
\includegraphics[width=\textwidth]{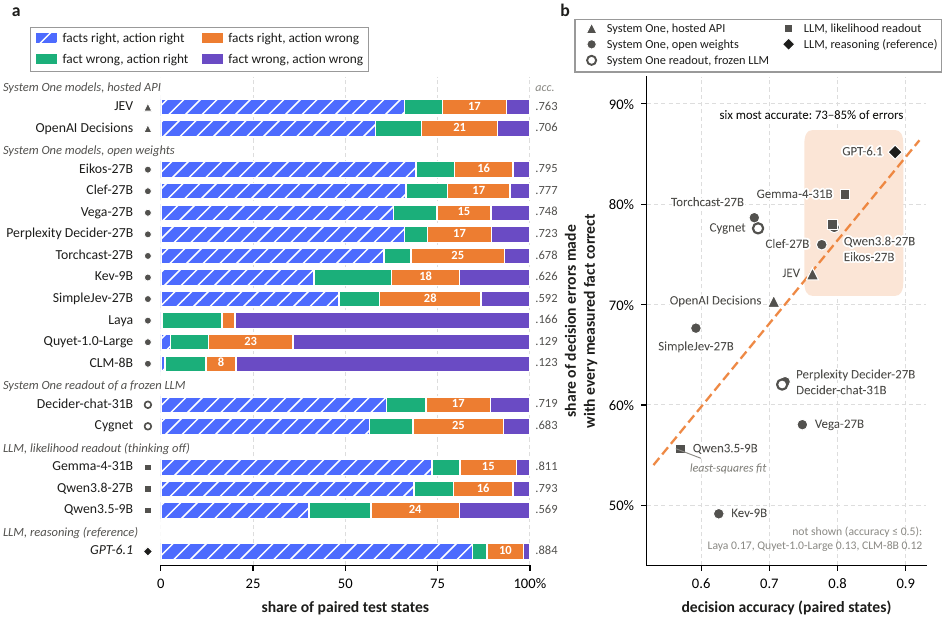}
\caption{Paired outcomes on the 19,663 paired test states. (a) Share of states in each outcome; the white number is the
share with every fact right but the wrong action. (b) Decision accuracy against the share of errors made with
every fact right; marker shape gives the system category. The more accurate the system, the larger this share.}
\label{fig:paired}
\end{figure*}

We first ask how often measured comprehension and action correctness come apart, how costly the
errors are, and whether they are idiosyncratic to a model or attached to particular states.

\subsection{Most decision errors follow correct reading}
\label{sec:dissociation}

Table~\ref{tab:paired} and Figure~\ref{fig:paired} give the four outcomes for all eighteen systems on the
same 19,663 states. For the six most accurate systems, 73--85\% of all decision errors occur on
states where every measured fact is answered correctly. Conditional on all measured facts being
correct, the action is still wrong in 10--21\% of states. The converse also holds: when at least one
fact is answered wrongly, the action is nevertheless correct in 62--70\% of states. Neither direction of
the pairing is a reliable proxy for the other.

The share of errors that follow correct reading grows with accuracy
(Figure~\ref{fig:paired}b). Among the fast systems above 0.5 accuracy it ranges from 49\% (Kev-9B) to 81\%
(Gemma), and it is 85\% for the reasoning reference, GPT-6.1. Stronger systems mainly stop misreading;
they do not stop choosing wrongly after reading correctly. Deliberation helps: GPT-6.1 raises decision
accuracy to 0.884, 7 points above Gemma and 9--12 points above the three most accurate System One models
(Eikos-27B, Clef and JEV), at about 70 times JEV's cost per state (\S\ref{sec:efficiency}). But its remaining
errors are of the same kind, and in an even larger proportion. CLM-8B, Laya and Quyet-1.0-Large rarely answer
both the probes and the action correctly (U1D1 1.1\%, 0.4\% and 2.7\%), largely because of input limits that we
count as wrong (\S\ref{sec:systems}).

The pattern does not depend on the type of system. Among the System One models, the hosted APIs make 73\% (JEV)
and 70\% (OpenAI Decisions) of their errors after correct reading, the open-weight models above 0.5 accuracy
from 49\% (Kev-9B) to 79\% (Torchcast-27B; 78\% for Eikos-27B, the most accurate System One model at 0.795), and the two readouts of a frozen LLM 62\%
(Decider-chat-31B) and 78\% (Cygnet); the LLM likelihood readouts make 56--81\%. Systems differ in how often a
correct reading still leads to the wrong action: OpenAI Decisions acts wrongly in 26\% of states with every fact
right, Torchcast-27B in 30\% and SimpleJev-27B in 36\%, against 17--21\% for the five most accurate fast systems; SimpleJev-27B, which
reads well ($U$ = 0.758) but decides correctly on only 0.592 of paired states, thus makes 68\% of its errors after
correct reading. Torchcast-27B is the clearest case: it reads better than any other System One model ($U$ =
0.858, second among fast systems only to Gemma) but decides correctly on only 0.678 of paired states, the
largest gap between reading and acting of any competent system, so that 79\% of its errors follow correct
reading; it is weakest on ALFWorld (0.565). Perplexity Decider-27B and Decider-chat-31B rarely act correctly after a misread fact
($P(D\,|\,U{=}0)$ 0.38 and 0.50), partly because a refused or shortened long state counts against both the probes
and the decision (on $\tau^2$-bench, with the longest states, they reach 0.53 and 0.48); Vega-27B, strong in
TextWorld (0.880), is weak on $\tau^2$-bench and on the visible-field probes (0.60 and 0.61), where its server fails on the
longest states.

\begin{figure*}[!tp]
\centering
\includegraphics[width=\textwidth]{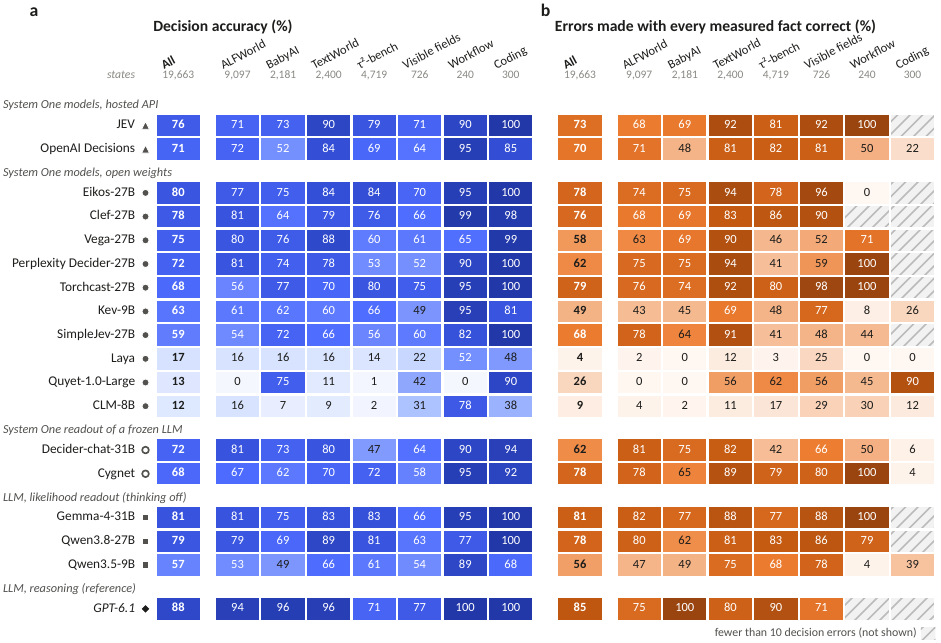}
\caption{(a) Decision accuracy and (b) share of errors made with every fact right, overall and by domain (number of
states under each label; ``Visible fields'': visible-field probes, \S\ref{app:offline}). Hatched: fewer than ten
errors.}
\label{fig:domains}
\end{figure*}

The pattern is not driven by one domain (Figure~\ref{fig:domains}). Among the six strongest
systems, the share of errors after correct reading is 68--82\% in ALFWorld, 62--100\% in BabyAI,
80--94\% in TextWorld and 77--90\% in $\tau^2$-bench. Two panels make the dissociation especially
visible. On a TextWorld cohort generated after all earlier runs, JEV and Gemma both answer every probe
correctly, yet their decision accuracies are 95.0\% and 73.2\%; Gemma has 161 of 600 states with all
facts right and the action wrong, against 30 for JEV. On coding traces, both choose the correct next
action on all 300 test states while answering the paired probes at 85.2\% and 97.1\%. A saturated action
score here hides a twelve-point difference in what the models read correctly. Rankings also change across
domains: among the fast systems, Decider-chat-31B, Gemma and Clef lead on ALFWorld (0.81--0.82), JEV on
TextWorld and on the visible-field probes, and Eikos-27B on $\tau^2$-bench (0.843), while GPT-6.1, strongest everywhere
else, is the weakest of the six strongest systems on $\tau^2$-bench (0.709).
The reason is a preference for caution rather than misreading. On $\tau^2$ states whose gold action is a
tool call, typically a write the user has just confirmed, GPT-6.1 replies to the user instead in 16.7\% of
cases, against 2.8--7.0\% for Gemma, Qwen3.8-27B, Clef and JEV, and calls a read-only tool to re-verify in another
5.3\%; together these account for 80\% of its errors there. Where the correct next step is to reply, it is
more accurate than these four (0.695 vs.\ 0.34--0.64). Because $\tau^2$ gold actions come from successful
trajectories, this extra caution is scored as an error.

\subsection{Errors after correct reading are cheaper, but not free}
\label{sec:severity}

\begin{figure*}[!tp]
\centering
\includegraphics[width=\textwidth]{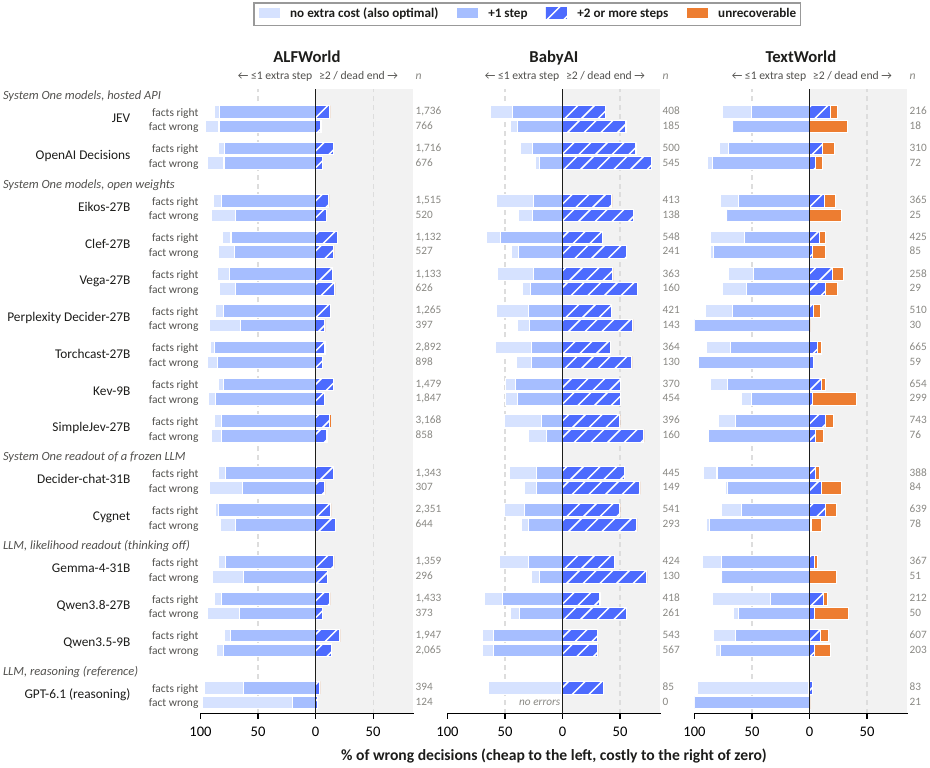}
\caption{Cost of every wrong decision in the three simulators, by exact lookahead (extra steps added to the shortest
remaining plan). Left of zero: at most one extra step; right: two or more, or unrecoverable. Upper bar: errors
with every fact right; lower bar: with a wrong fact; $n$: wrong decisions. Systems below 0.5 accuracy or
failing the coverage rule are omitted.}
\label{fig:severity}
\end{figure*}

A wrong action can be a harmless detour or a dead end. In the three simulators we replay every wrong
decision and compute its regret by exact lookahead (Figure~\ref{fig:severity}). Pooled over the 14 competent
fast systems (paired decision accuracy above 0.5, coverage rule passed), most errors cost at most one extra step:
86--90\% in ALFWorld, 42--56\% in BabyAI and 77--84\% in TextWorld, the range spanning errors after correct
reading and after a misread fact. Where the two kinds of error differ, errors after a misread fact are the
costlier ones. In BabyAI, 58\% of them cost two or more steps, against 44\% of errors after correct reading; in
TextWorld, 20\% are unrecoverable, against 6\%. For Gemma in TextWorld the unrecoverable share is 24\% after a
misread fact and 2\% otherwise. ALFWorld is the exception: almost nothing is unrecoverable there, and errors after
correct reading are, if anything, slightly costlier (14\% vs.\ 10\% cost two or more extra steps). Between 5\%
and 42\% of a system's ``wrong'' decisions, depending on the system and the simulator, are in fact also optimal
under exact lookahead, which bounds how much of the measured error comes from gold sets that are stricter than
necessary.

The reasoning reference changes this picture. GPT-6.1 makes few errors after a misread fact in the
simulators (124 in ALFWorld, none in BabyAI, 21 in TextWorld), and a large part of its errors after
correct reading are not errors at all: exact lookahead shows that the chosen action is also optimal for
34\% of them in ALFWorld, 64\% in BabyAI and 96\% in TextWorld. As decision accuracy rises, gold sets
that admit only some of the optimal actions account for a growing share of the measured error, and the
headline share of errors after correct reading (85\% for GPT-6.1) includes many such cases. Exact
lookahead is therefore a necessary complement to set-valued gold when strong models are compared.

\subsection{The same states defeat many systems}
\label{sec:consistency}

\begin{figure*}[!tp]
\centering
\includegraphics[width=\textwidth]{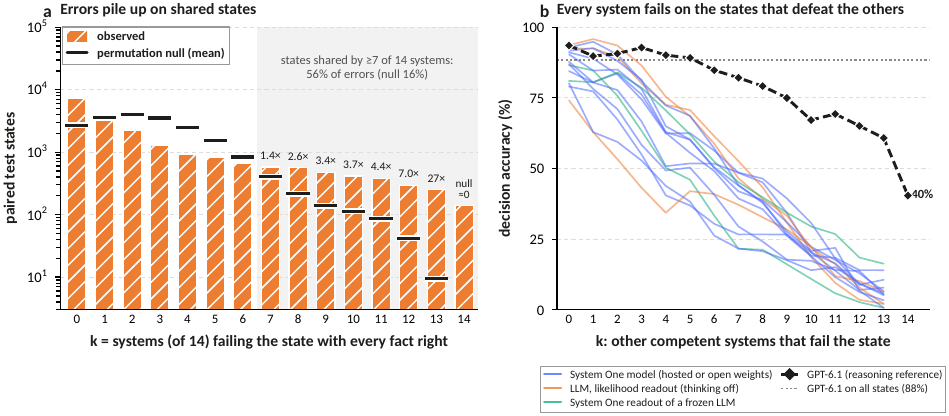}
\caption{(a) Paired test states on which $k$ of the 14 competent fast systems make a facts-right, action-wrong error,
against a within-cell permutation null (ticks; labels give observed/null). (b) Accuracy of each competent system
on states that $k$ of the other thirteen fail (thin lines, by category) and of GPT-6.1 by $k$ (dashed).}
\label{fig:consistency}
\end{figure*}

If facts-right, action-wrong errors were idiosyncratic noise, they would be spread evenly over states.
They are not (Figure~\ref{fig:consistency}). Of the 53,382 such events produced by the 14 competent fast
systems, 56\% fall on states where at least seven of the 14 fail in this way, against 16\% under a null that
permutes each system's events within its frozen cells, and 85\% on states shared by at least three (78\% under
the null). All 14 systems fail with every fact right on the same 141 states, where the null expects none, and each
of them decides correctly on 74--94\% of the states that no other competent system fails but on only 1--16\% of
those that all thirteen others fail (Figure~\ref{fig:consistency}b). The
errors are therefore attached to properties of particular states, which makes them worth explaining
(\S\ref{sec:mechanism}); and a benchmark that samples such states is a sharper diagnostic than one that weights
all states equally. Deliberation resolves many of them but not all: GPT-6.1, accurate on 88\% of all paired
states, chooses correctly on only 40\% of the 141 states that defeat every competent fast system, and 62\% of its
own facts-right, action-wrong errors fall on states where at least three of them also fail.

\subsection{How much of this is label error?}
\label{sec:labels}

A facts-right, action-wrong outcome may also come from a wrong label. Two audits bound this. An
independent content audit found 42 test records with conflicting labels and 56 whose inputs also occur
in dev; excluding these 98 records changes no group score of any of the eighteen systems by more than 3.5 points
(\S\ref{app:freeze}). Second, while building the decision-oriented probes of \S\ref{sec:dprobes} we
re-derived every ALFWorld label from the text of the trace alone. In the original ALFWorld cohorts,
several receptacles can share one simulator location, and the builder recorded only one of them. For
303 paired test records, the location label therefore names a co-located receptacle rather than the one
the text says the agent arrived at, and the four strong systems we audited answer the receptacle named in the
text (98--99\%), so these records cannot reach $U{=}1$ (Appendix~\ref{app:audit}). The later confirmation
cohort uses set-valued labels and is unaffected. Excluding the 398 affected ALFWorld records raises the
share of errors made after correct reading for every audited system (on ALFWorld, from 0.663 to 0.707 for JEV
and from 0.810 to 0.885 for Gemma). We keep the frozen labels and report this as a
sensitivity analysis. It makes our headline estimate conservative.

%% file: tables/paired.tex
\begin{table*}[!tp]
\centering
\small
\definecolor{tabmuted}{HTML}{8a8984}\definecolor{tabtint}{HTML}{FBE5D6}
\setlength{\tabcolsep}{5pt}
\begin{tabular}{@{}lrrrrrrrrrr@{}}
\toprule
 & \multicolumn{2}{c}{Accuracy} & \multicolumn{4}{c}{Joint outcome (\% of states)} & \multicolumn{2}{c}{$P(D{=}1\mid U)$} & \begin{tabular}[b]{@{}r@{}}Errors with\end{tabular} & \begin{tabular}[b]{@{}r@{}}Coverage\end{tabular} \\
\cmidrule(lr){2-3}\cmidrule(lr){4-7}\cmidrule(lr){8-9}
System & $D$ & $U$ & U1D1 & \cellcolor{tabtint}U1D0 & U0D1 & U0D0 & $U{=}1$ & $U{=}0$ & $U{=}1$ (\%) & (\%) \\
\midrule
\multicolumn{11}{@{}l}{\emph{System One models, hosted API}} \\
JEV & 0.763 & 0.833 & 66.0 & \cellcolor{tabtint}17.3 & 10.4 & 6.4 & 0.792 & 0.619 & 73.1 & 100.0 \\
OpenAI Decisions & 0.706 & 0.789 & 58.2 & \cellcolor{tabtint}20.6 & 12.4 & 8.7 & 0.738 & 0.588 & 70.3 & 99.6 \\
\midrule
\multicolumn{11}{@{}l}{\emph{System One models, open weights}} \\
Eikos-27B & 0.795 & 0.850 & 69.1 & \cellcolor{tabtint}15.9 & 10.4 & 4.6 & 0.813 & 0.695 & 77.7 & 100.0 \\
Clef-27B & 0.777 & 0.833 & 66.4 & \cellcolor{tabtint}17.0 & 11.3 & 5.4 & 0.796 & 0.678 & 76.0 & 100.0 \\
Vega-27B & 0.748 & 0.776 & 63.0 & \cellcolor{tabtint}14.6 & 11.9 & 10.6 & 0.812 & 0.529 & 58.0 & 95.1 \\
Perplexity Decider-27B$^\dagger$ & 0.723 & 0.832 & 66.0 & \cellcolor{tabtint}17.3 & 6.3 & 10.4 & 0.793 & 0.378 & 62.3 & 90.7 \\
Torchcast-27B & 0.678 & 0.858 & 60.4 & \cellcolor{tabtint}25.3 & 7.3 & 6.9 & 0.705 & 0.517 & 78.7 & 100.0 \\
Kev-9B & 0.626 & 0.600 & 41.6 & \cellcolor{tabtint}18.4 & 21.0 & 19.0 & 0.693 & 0.525 & 49.2 & 100.0 \\
SimpleJev-27B$^\dagger$ & 0.592 & 0.758 & 48.2 & \cellcolor{tabtint}27.6 & 11.0 & 13.2 & 0.636 & 0.454 & 67.6 & 93.4 \\
Laya & 0.166 & 0.039 & 0.4 & \cellcolor{tabtint}3.5 & 16.2 & 79.9 & 0.110 & 0.168 & 4.1 & 97.0 \\
Quyet-1.0-Large$^\dagger$ & 0.129 & 0.256 & 2.7 & \cellcolor{tabtint}23.0 & 10.2 & 64.2 & 0.104 & 0.137 & 26.3 & 3.8 \\
CLM-8B & 0.123 & 0.092 & 1.1 & \cellcolor{tabtint}8.1 & 11.2 & 79.6 & 0.122 & 0.123 & 9.2 & 100.0 \\
\midrule
\multicolumn{11}{@{}l}{\emph{System One readout of a frozen LLM}} \\
Decider-chat-31B & 0.719 & 0.786 & 61.2 & \cellcolor{tabtint}17.4 & 10.7 & 10.7 & 0.778 & 0.501 & 62.0 & 99.7 \\
Cygnet & 0.683 & 0.812 & 56.6 & \cellcolor{tabtint}24.6 & 11.7 & 7.1 & 0.697 & 0.623 & 77.6 & 99.7 \\
\midrule
\multicolumn{11}{@{}l}{\emph{LLM, likelihood readout (thinking off)}} \\
Gemma-4-31B & \textbf{0.811} & \textbf{0.888} & 73.5 & \cellcolor{tabtint}15.3 & 7.6 & 3.6 & \textbf{0.828} & 0.678 & 80.9 & 100.0 \\
Qwen3.8-27B & 0.793 & 0.848 & 68.6 & \cellcolor{tabtint}16.2 & 10.6 & 4.6 & 0.809 & \textbf{0.699} & 78.0 & 100.0 \\
Qwen3.5-9B & 0.569 & 0.640 & 40.1 & \cellcolor{tabtint}23.9 & 16.8 & 19.1 & 0.626 & 0.468 & 55.6 & 100.0 \\
\midrule
\multicolumn{11}{@{}l}{\emph{LLM, reasoning (reference)}} \\
GPT-6.1$^\ast$ & 0.884 & 0.943 & 84.4 & \cellcolor{tabtint}9.9 & 4.0 & 1.7 & 0.895 & 0.701 & 85.2 & 99.2 \\
\bottomrule
\end{tabular}
\caption{Paired outcomes on the 19,663 paired test states (pooled counts). $D$: decision accuracy; $U$: share of states with every measured fact correct. The last two columns give the share of errors made with $U{=}1$ and coverage (states with every answer supported). Bold: best fast system. $^\ast$Reasoning reference. $^\dagger$Coverage below 95\%.}
\label{tab:paired}
\end{table*}

%% file: sections/07_mechanism.tex
\section{Why Do Correct Readers Decide Wrongly?}
\label{sec:mechanism}

\begin{figure*}[!tp]
\centering
\includegraphics[width=\textwidth]{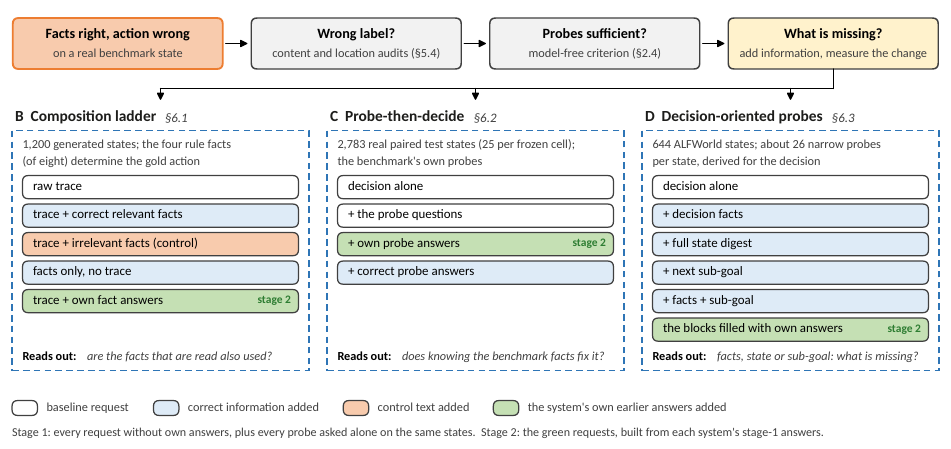}
\caption{The diagnostic studies. Each sends the decision again with information added, as separate requests on the
same states: B on generated states whose probes determine the action, C with the benchmark's probes on real
states, D with probes derived for the decision on ALFWorld. Green requests use the system's own earlier answers.}
\label{fig:design}
\end{figure*}

A facts-right, action-wrong outcome on a natural state has two readings. The probes may have missed a
decision-relevant fact (\S\ref{sec:sufficiency}), or the model may have read everything it needed and
still not combined it into the right action. On natural states the two cannot be separated, so we first
build states where they can, and then return to real benchmark states.
Figure~\ref{fig:design} summarizes the three studies. Each adds information to the decision request and
measures the change on the same states; requests that contain a system's own answers are built from its
earlier answers to the same probes, asked alone.

\subsection{When the probes are sufficient by construction}
\label{sec:ladder}

\paragraph{Design.} We generate decision states in two domains, a household text world and a customer
service policy. Each state has a goal, a numbered decision list of four
rules followed by a default action (``apply the first rule whose condition holds''), an executed trace that
establishes eight binary facts, and a fixed set of nine actions. Four of the facts are tested by the rules;
the other four are distractors, each tied to a plausible action outside the procedure, and any fact may be
established and then overwritten by later feedback. The ladder variable is the position of the first rule
whose condition holds (1--4, or none, so that the default applies). Later rules hold independently with
probability 0.5, so that choosing a later rule whose condition also holds (a precedence error) can be told
apart from choosing a rule whose condition does not hold. For every assignment of the four rule facts the
gold action is unique, which we verify by enumeration. The probes ask the four rule facts and two of the
distractors, so correct answers to the rule-fact probes determine the action, and $U{=}1,D{=}0$ can only be
a failure to use facts that were read. On the same 1,200 test states (120 per domain and ladder position) we
ask the decision under four inputs: the raw trace; the trace plus the current values of the four rule facts;
the trace plus the current values of the four distractor facts, in the same form (a control for added
text); and the values of all eight facts without the trace.

\begin{figure*}[!tp]
\centering
\includegraphics[width=\textwidth]{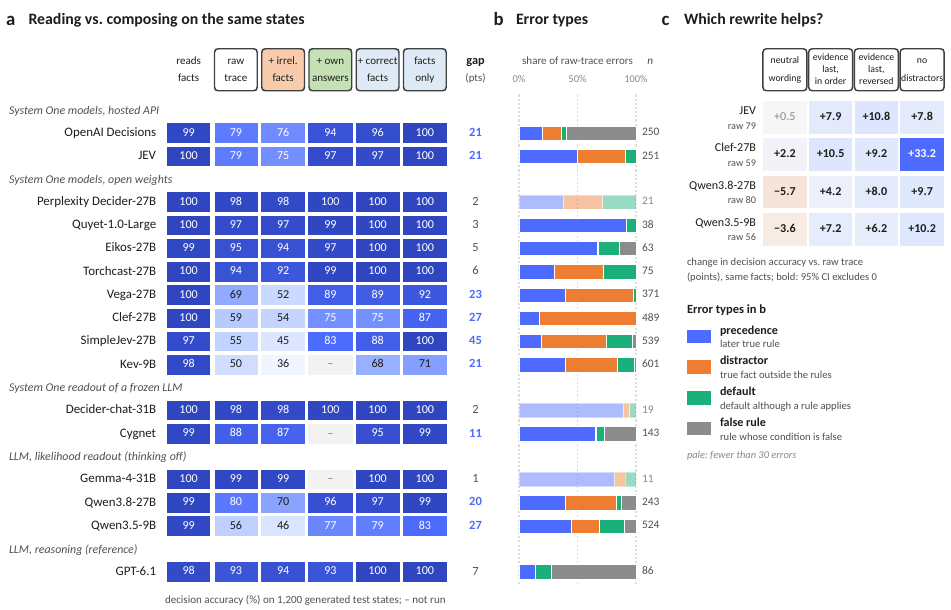}
\caption{States whose probes are sufficient by construction (1,200 test states). (a) Probe accuracy and decision
accuracy under each input, and the gap from raw trace to facts only (points). (b) Raw-trace errors by type ($n$:
errors). (c) Change vs.\ the raw trace for rewrites with the same facts; bold: 95\% interval excludes zero. Header
colours follow Figure~\ref{fig:design}.}
\label{fig:ladder}
\end{figure*}

\paragraph{Reading and composing each succeed; doing both at once does not.} Every system shown in
Figure~\ref{fig:ladder}a answers the rule-fact probes almost perfectly on the raw trace (97--100\%). JEV and
Qwen3.8-27B also decide perfectly when given the facts alone (1.000 and 0.994). Yet from the raw trace
their decision accuracy is 0.791 and 0.797, a gap of 20 points between two abilities that each work in
isolation. Appending the correct values of the relevant facts to the trace closes the gap (0.974 and
0.967), while appending the same amount of irrelevant text widens it (0.751 and 0.705). Gemma shows no gap
at all (0.991 from the trace), and Cygnet lies in between, with 0.992 from the facts alone and 0.881 from
the trace. For Clef, Qwen3.5-9B and Kev-9B even the facts-only condition is imperfect
(0.71--0.87), so they also fail to apply the rules. For the fast systems, almost none of the errors comes
from not knowing a fact (97--100\% of JEV's, Qwen3.8-27B's and Cygnet's errors have all four rule facts right).
They come from choosing what to act on. The reasoning reference is the exception: GPT-6.1 decides at 0.928
from the trace and at 1.000 from the facts, but only 26 of its 86 errors have all four rule facts right, and 62
apply a rule whose condition does not hold. When it reasons before answering, composition is no longer the
bottleneck and the remaining errors are misreadings, spread evenly over facts that were and were not
overwritten later in the trace (2.6\% and 2.3\% of probes).

The remaining eight System One models separate the two abilities just as cleanly: seven of them decide
perfectly from the facts alone (1.000). Four of them also nearly close the gap from
the raw trace: Decider-chat-31B, Perplexity Decider-27B, Quyet-1.0-Large and Eikos-27B decide at
0.95--0.98, and most of their few errors (19--63) are precedence errors with all facts right. Torchcast-27B comes
close (0.938 from the trace, a gap of 6 points); its 75 errors, all but one with every fact right, are spread
over precedence, distractor and default actions. OpenAI
Decisions repeats the pattern of JEV and Qwen3.8-27B: 0.792 from the trace, 0.964 with the relevant facts
appended and 0.757 with irrelevant ones; 232 of its 250 errors have all four rule facts right. A gap of 20
points between reading and composing is therefore not a property of one model family, but neither is it
universal among fast decision models. Vega-27B shows a gap of its own: it reads the facts (0.999) but
decides at 0.691 from the trace, 0.918 from the facts alone and 0.520 when irrelevant facts are appended,
and 216 of its 371 errors act on a true fact outside the procedure, the capture failure we see for Clef.
SimpleJev-27B has the largest gap of any system: it decides perfectly from the facts alone but at 0.551 from
the trace and 0.451 with irrelevant facts appended, and 299 of its 539 errors act on a fact outside the
procedure.

\paragraph{The errors follow true but wrong cues.} JEV never applies a rule whose condition is false
(OpenAI Decisions does, in 149 of its 250 errors, at least 131 of them with all four rule facts answered correctly). Its
251 errors on the raw trace apply a later rule whose condition is also true, skipping an earlier one (125,
\emph{precedence}), act on a true fact that is outside the procedure (103, \emph{distractor}), or fall back
to the default action although a rule applies (23). Across systems (Figure~\ref{fig:ladder}b), the errors of Clef,
Vega-27B and SimpleJev-27B are mostly distractor actions, the few errors of Quyet-1.0-Large, Eikos-27B and
Decider-chat-31B mostly precedence errors, and those of GPT-6.1 and OpenAI Decisions mostly false rules. We test four
explanations with targeted rewrites of the same states (Figure~\ref{fig:ladder}c). Rewriting observations as neutral statements changes nothing for JEV (+0.5),
so the wording of observations is not the cause. Moving the decisive evidence to the end of the trace
helps (+7.9 in rule order, +10.8 reversed), and removing the distractor facts also helps (+7.8), but
neither removes the precedence errors. Only a summary of the facts in one place recovers the facts-only
level. For JEV and Qwen3.8-27B, the failure is therefore one of \emph{consolidating} evidence scattered over
a trace before applying an ordered rule. Clef behaves differently: removing distractor facts alone raises
its accuracy by 33 points, to 0.924, so its dominant failure is \emph{capture} by out-of-procedure cues.
The same benchmark thus separates two mechanisms that yield the same outcome.

\paragraph{Using the model's own answers.} Because these models read the facts correctly, they can supply
the summary themselves. Appending each system's own probe answers (asked in separate requests) recovers
almost all of the gap: JEV rises from 0.791 to 0.968 (+17.7), Qwen3.8-27B to 0.955 (+15.8), Qwen3.5-9B to
0.767 (+20.3) and Clef to 0.747 (+15.4; its self-written summaries are identical to the correct ones on
1,197 of 1,200 states). GPT-6.1 gains nothing (0.928 to 0.931): its errors are misreadings, and its own
summaries repeat them (114 wrong fact values over the 1,200 states). OpenAI Decisions recovers most of its
gap from its own answers (0.792 to 0.937, +14.5, CI [+12.1, +16.8]), although its summaries contain 63 wrong
fact values; Vega-27B gains about as much as Qwen3.5-9B (0.691 to 0.890, +19.9, CI [+17.7, +22.2]) with only 5
wrong fact values; SimpleJev-27B gains the most of all (0.551 to 0.827, +27.6, CI [+25.1, +30.2]) despite
148 wrong fact values; Torchcast-27B closes its smaller gap (0.938 to 0.994, +5.7, CI [+4.3, +7.0]); the
four systems near ceiling gain 1.3--2.7 points. Composing the action in code from the model's own fact answers, which is possible
here because the rules are explicit, is better still: 0.996 for JEV, 0.987 for Qwen3.8-27B, 1.000 for Clef
and 0.965 for Qwen3.5-9B. The second request can be reserved for uncertain states: using the native
confidence of the first decision, a cascade that re-asks only below a confidence of 0.7 reaches 0.932 for
JEV with a second request on 38\% of states. This is the usage pattern that System One models are designed
for: many narrow questions in parallel, combined by code \citep{typesafe2026jev}.

\subsection{Back to real states: probe-then-decide}
\label{sec:ptd}

\input{tables/probe_then_decide}

\begin{figure*}[!tp]
\centering
\includegraphics[width=\textwidth]{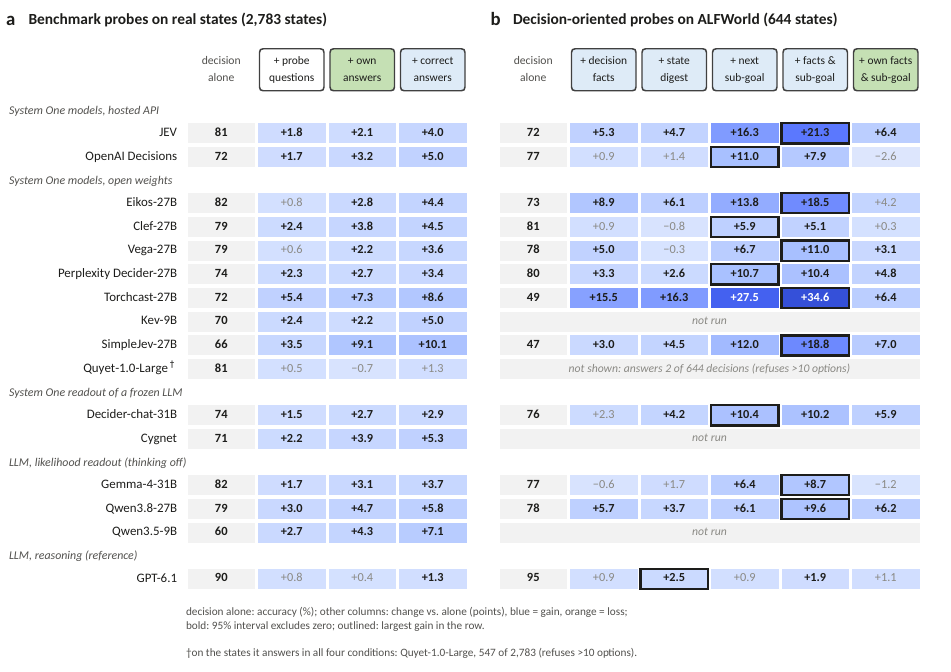}
\caption{Adding information to real states: decision accuracy alone (gray) and the change when information is added
(points; blue gain, orange loss; bold: 95\% interval excludes zero). (a) Benchmark probes on 2,783 paired states.
(b) Decision-oriented blocks on 644 ALFWorld states; outlined: the largest gain in each row. Header colours follow
Figure~\ref{fig:design}.}
\label{fig:interventions}
\end{figure*}

Does the remedy transfer to natural states? We sample 2,783 paired test states (25 per frozen cell, all
domains) and ask the decision alone, after the benchmark's probe questions, after the system's own earlier
probe answers, or after the correct answers (Table~\ref{tab:ptd}, Figure~\ref{fig:interventions}a). The
effect is small for every system but two. For the other fast systems, correct answers add 2.9--7.1 points, the
system's own answers 2.1--4.7 and merely asking the questions 0.6--3.0; GPT-6.1, which decides alone at 0.902, gains 1.3 points
from the correct answers and 0.4 from its own. Most of the benefit of the own answers is therefore already
obtained by listing the questions: the answers themselves add at most 2.1 points, and nothing distinguishable
from zero for JEV, Kev-9B, Qwen3.5-9B and Perplexity Decider-27B. SimpleJev-27B and Torchcast-27B are the exceptions. Deciding
alone at 0.656, SimpleJev-27B gains 10.1 points from the correct answers and 9.1 from its own, 5.6 of them
beyond the questions; Torchcast-27B, alone at 0.715, gains 8.6 from the correct answers and 7.3 from its own,
but 5.4 of them already from the questions, mostly on ALFWorld (correct answers +20.2 there).
The small net gain hides larger movements. For Qwen3.8-27B, JEV, Gemma and Clef, the own answers
repair 34\%, 32\%, 27\% and 25\% of the facts-right, action-wrong states, but break 2--4\% of the states that
were correct, which are about five times as many. OpenAI Decisions' own answers repair 33\% of its
facts-right, action-wrong states, as many as for JEV, but break 6.8\% of the correct ones, the most of any
system. Quyet-1.0-Large answers only 547 of the 2,783 states (it refuses option lists longer than ten); on
those, no condition changes its accuracy measurably (0.808 alone).

The contrast with \S\ref{sec:ladder} is the point. There, the probes were sufficient and their answers
determined the action; here, even the correct answers to the benchmark probes leave most errors in place.
The probes were designed to measure reading, not to specify the decision, so they do not carry the
missing information.

\subsection{Decision-oriented probes}
\label{sec:dprobes}

\input{tables/decision_probes}

If the bottleneck is what the probes ask, probes designed for the decision should do better. On 644
ALFWorld states we derive, from the text of the trace alone, everything a planner would need: the agent's
location and inventory, where each still-needed object was last seen, whether the held object has been
cleaned, heated or cooled, how many objects are already in place, which receptacles have not yet been
visited, and the next sub-goal (search, fetch, take, go to the appliance, process, go to the destination,
place, use the lamp). We validate the parse against the simulator labels (\S\ref{sec:labels}) and ask each
fact as a narrow, independent question, about 26 per state. We then add to the decision request either the
decision-oriented facts, a complete digest of the observed state, the next sub-goal, or facts plus
sub-goal, each with correct values, and the same blocks filled with each system's own answers
(Table~\ref{tab:dprobes}, Figure~\ref{fig:interventions}b).

\paragraph{The facts are read; the sub-goal is not chosen.} JEV answers the narrow probes almost
perfectly: 99.7\% of the 14,104 visited-or-not questions, 100\% for the held object and placed count,
96.9\% for where needed objects were seen. Yet decision-oriented facts add only 5.3 points, no more than the
benchmark's own probes (5.7), and a complete state digest adds 4.7. Stating the next sub-goal adds 16.3
points, and facts plus sub-goal 21.3, reaching 0.929. JEV's own answer to the sub-goal question is right in
only 66.8\% of states, so the same block filled with its own answers adds 6.4. Of JEV's 183 errors when
deciding alone, 41\% have every fact right but the sub-goal wrong, and another 37\% have both right when
asked separately: JEV can state the sub-goal it then fails to pursue. For JEV, the remaining errors on real
ALFWorld states are errors of planning, not of reading or of consolidation.

\paragraph{Stronger systems are limited elsewhere.} The same blocks help Clef, Gemma and Qwen3.8-27B much
less: no block adds more than 10 points, and the sub-goal adds about 6. Their own answers help only
Qwen3.8-27B (+4.8 with facts, +6.2 with facts and sub-goal); for Clef and Gemma they change nothing
($-1.2$ to $+0.3$). GPT-6.1 decides alone at 0.953 and no block, correct or its own, adds more than 2.5
points. How well a system answers the sub-goal question does not predict whether its answers
help. Clef answers it best (89.1\%) and gains nothing; of Clef's errors when deciding alone, nearly half
(48\%) occur with every fact and the sub-goal answered correctly in separate requests. Qwen3.8-27B answers
it worst (36.0\%, below its own decision accuracy) and still gains, through the facts; GPT-6.1 answers it
equally poorly (35.2\%) while deciding correctly in 95\% of states, so its decisions do not pass through
the sub-goal as the benchmark labels it. The decision-oriented
facts can even
hurt: on states where one needed object has been seen and another has not, listing the unvisited
receptacles sends Gemma and Qwen3.8-27B to search instead of fetching what they already know about, and
Gemma's accuracy on these states falls from 0.659 to 0.341. With every block given, most of Gemma's
remaining errors pick the right kind of action for the wrong object. Extra information is not neutral: a
salient but incomplete summary can pull a model away from the trace it otherwise reads well.

\paragraph{The other systems.} The seven other System One models that answer ALFWorld decisions confirm that the
sub-goal is the most useful single block: it adds 6.7--13.8 points for six of them (OpenAI Decisions +11.0, CI
[+7.2, +14.9]; Eikos-27B +13.8), while decision-oriented facts add 0.9--3.3 points, except for Vega-27B (+5.0) and
Eikos-27B (+8.9), which gains 18.5 from facts plus sub-goal. Torchcast-27B repeats JEV's pattern more strongly:
deciding alone at only 0.486, it gains 27.5 points from the sub-goal and 34.6 from facts plus sub-goal (to 0.832),
more than any other system, answers the sub-goal question correctly in only 60\% of states, and 53\% of its errors
when deciding alone have every fact right but the sub-goal wrong. Like Clef and Gemma rather
than JEV, the others mostly answer the sub-goal question well (0.77--0.87; SimpleJev-27B 0.64), and 32--64\% of their errors when deciding
alone occur with every fact and the sub-goal answered correctly in separate requests. Their own answers help less than the
correct ones, and OpenAI Decisions' own facts and sub-goal even lower its accuracy ($-2.6$, CI [$-7.1$,
$+1.9$]). The pull of an incomplete summary also recurs: on the states where one needed object has been
seen and another has not, correct decision facts lower OpenAI Decisions from 0.698 to 0.543 and
Decider-chat-31B from 0.698 to 0.434, although Vega-27B and Torchcast-27B gain on the same states (0.713 to 0.806 and 0.302 to 0.496).

\paragraph{What this means.} Across these studies, the source of a facts-right, action-wrong error
varies by system and by state: consolidation of scattered evidence for JEV and Qwen3.8-27B on rule
procedures, capture by irrelevant cues for Clef, sub-goal selection for JEV on household tasks, and
grounding a sub-goal in the right object for Gemma. Each calls for a different remedy, and only the paired
measurement, combined with probes that are sufficient for the decision, reveals which applies.

%% file: tables/probe_then_decide.tex
\begin{table}[t]
\centering
\small
\definecolor{tabmuted}{HTML}{8a8984}\definecolor{tabtint}{HTML}{FBE5D6}
\setlength{\tabcolsep}{2.4pt}
\begin{tabular}{@{}lrrrr@{}}
\toprule
 & & \multicolumn{3}{c}{Change vs.\ alone (points)} \\
\cmidrule(l){3-5}
System & Alone & +questions & +own & +correct \\
\midrule
\multicolumn{5}{@{}l}{\emph{System One models, hosted API}} \\
JEV & 0.806 & +1.8 & +2.1 & +4.0 \\
 & & {\scriptsize [+0.4, +3.2]} & {\scriptsize [+0.7, +3.3]} & {\scriptsize [+2.6, +5.5]} \\
OpenAI Decisions & 0.722 & +1.7 & +3.2 & +5.0 \\
 & & {\scriptsize [+0.0, +3.5]} & {\scriptsize [+1.6, +5.0]} & {\scriptsize [+3.4, +6.7]} \\
\midrule
\multicolumn{5}{@{}l}{\emph{System One models, open weights}} \\
Eikos-27B & \textbf{0.821} & \textcolor{tabmuted}{+0.8} & +2.8 & +4.4 \\
 & & {\scriptsize [\textminus{}0.3, +1.8]} & {\scriptsize [+1.6, +4.1]} & {\scriptsize [+3.1, +5.8]} \\
Clef-27B & 0.789 & +2.4 & +3.8 & +4.5 \\
 & & {\scriptsize [+1.1, +3.6]} & {\scriptsize [+2.6, +5.2]} & {\scriptsize [+3.1, +5.9]} \\
Vega-27B & 0.792 & \textcolor{tabmuted}{+0.6} & +2.2 & +3.6 \\
 & & {\scriptsize [\textminus{}0.5, +1.6]} & {\scriptsize [+1.0, +3.4]} & {\scriptsize [+2.3, +4.8]} \\
Perplexity & 0.739 & +2.3 & +2.7 & +3.4 \\
Decider-27B & & {\scriptsize [+1.4, +3.4]} & {\scriptsize [+1.7, +3.8]} & {\scriptsize [+2.4, +4.6]} \\
Torchcast-27B & 0.715 & +5.4 & +7.3 & +8.6 \\
 & & {\scriptsize [+4.2, +6.7]} & {\scriptsize [+5.8, +8.7]} & {\scriptsize [+7.1, +10.3]} \\
Kev-9B & 0.697 & +2.4 & +2.2 & +5.0 \\
 & & {\scriptsize [+1.2, +3.7]} & {\scriptsize [+0.8, +3.5]} & {\scriptsize [+3.6, +6.5]} \\
SimpleJev-27B & 0.656 & +3.5 & +9.1 & +10.1 \\
 & & {\scriptsize [+1.2, +5.8]} & {\scriptsize [+6.9, +11.4]} & {\scriptsize [+7.8, +12.5]} \\
Quyet-1.0-Large$^\dagger$ & 0.808 & \textcolor{tabmuted}{+0.5} & \textcolor{tabmuted}{\textminus{}0.7} & \textcolor{tabmuted}{+1.3} \\
 & & {\scriptsize [\textminus{}1.5, +2.4]} & {\scriptsize [\textminus{}3.1, +1.6]} & {\scriptsize [\textminus{}1.1, +3.7]} \\
\midrule
\multicolumn{5}{@{}l}{\emph{System One readout of a frozen LLM}} \\
Decider-chat-31B & 0.737 & +1.5 & +2.7 & +2.9 \\
 & & {\scriptsize [+0.7, +2.3]} & {\scriptsize [+1.6, +3.7]} & {\scriptsize [+1.9, +3.9]} \\
Cygnet & 0.713 & +2.2 & +3.9 & +5.3 \\
 & & {\scriptsize [+1.1, +3.4]} & {\scriptsize [+2.5, +5.6]} & {\scriptsize [+3.7, +6.8]} \\
\midrule
\multicolumn{5}{@{}l}{\emph{LLM, likelihood readout (thinking off)}} \\
Gemma-4-31B & \textbf{0.821} & +1.7 & +3.1 & +3.7 \\
 & & {\scriptsize [+0.9, +2.7]} & {\scriptsize [+2.0, +4.2]} & {\scriptsize [+2.5, +4.9]} \\
Qwen3.8-27B & 0.794 & +3.0 & +4.7 & +5.8 \\
 & & {\scriptsize [+1.3, +4.6]} & {\scriptsize [+3.2, +6.3]} & {\scriptsize [+4.5, +7.4]} \\
Qwen3.5-9B & 0.600 & +2.7 & +4.3 & +7.1 \\
 & & {\scriptsize [+1.1, +4.5]} & {\scriptsize [+1.6, +6.9]} & {\scriptsize [+4.4, +9.7]} \\
\midrule
\multicolumn{5}{@{}l}{\emph{LLM, reasoning (reference)}} \\
GPT-6.1 & 0.902 & \textcolor{tabmuted}{+0.8} & \textcolor{tabmuted}{+0.4} & +1.3 \\
 & & {\scriptsize [\textminus{}0.0, +1.7]} & {\scriptsize [\textminus{}0.1, +0.9]} & {\scriptsize [+0.7, +1.9]} \\
\bottomrule
\end{tabular}
\caption{Probe-then-decide on 2,783 paired test states: accuracy alone and change (points) after the probe questions, the system's own answers or the correct answers, with 95\% intervals below; gray: interval includes zero. Refused requests count as wrong (Perplexity Decider-27B: 268 states, 8k-token limit per question). $^\dagger$Quyet-1.0-Large: the 547 states it answers.}
\label{tab:ptd}
\end{table}

%% file: tables/decision_probes.tex
\begin{table*}[!tp]
\centering
\small
\definecolor{tabmuted}{HTML}{8a8984}\definecolor{tabtint}{HTML}{FBE5D6}
\setlength{\tabcolsep}{4pt}
\begin{tabular}{@{}lrrrrrrrrr@{}}
\toprule
 & & \multicolumn{7}{c}{Change vs.\ decision alone (points)} & \\
\cmidrule(lr){3-9}
 & & \multicolumn{5}{c}{Status block with correct values} & \multicolumn{2}{c}{With own answers} & \\
\cmidrule(lr){3-7}\cmidrule(lr){8-9}
System & \begin{tabular}[b]{@{}r@{}}Decision\\alone\end{tabular} & \begin{tabular}[b]{@{}r@{}}Benchmark\\probes\end{tabular} & \begin{tabular}[b]{@{}r@{}}Decision\\facts\end{tabular} & \begin{tabular}[b]{@{}r@{}}State\\digest\end{tabular} & \begin{tabular}[b]{@{}r@{}}Next\\sub-goal\end{tabular} & \begin{tabular}[b]{@{}r@{}}Facts +\\sub-goal\end{tabular} & \begin{tabular}[b]{@{}r@{}}\\Facts\end{tabular} & \begin{tabular}[b]{@{}r@{}}Facts +\\sub-goal\end{tabular} & \begin{tabular}[b]{@{}r@{}}Own sub-goal\\accuracy\end{tabular} \\
\midrule
\multicolumn{10}{@{}l}{\emph{System One models, hosted API}} \\
JEV & 0.716 & +5.7 & +5.3 & +4.7 & +16.3 & +21.3 & +4.8 & +6.4 & 0.668 \\
OpenAI Decisions & 0.767 & +4.0 & \textcolor{tabmuted}{+0.9} & \textcolor{tabmuted}{+1.4} & +11.0 & +7.9 & \textcolor{tabmuted}{+1.1} & \textcolor{tabmuted}{\textminus{}2.6} & 0.775 \\
\midrule
\multicolumn{10}{@{}l}{\emph{System One models, open weights}} \\
Eikos-27B & 0.725 & +9.3 & +8.9 & +6.1 & +13.8 & +18.5 & +6.8 & \textcolor{tabmuted}{+4.2} & 0.809 \\
Clef-27B & \textbf{0.806} & +4.8 & \textcolor{tabmuted}{+0.9} & \textcolor{tabmuted}{\textminus{}0.8} & +5.9 & +5.1 & \textcolor{tabmuted}{\textminus{}1.2} & \textcolor{tabmuted}{+0.3} & \textbf{0.891} \\
Vega-27B & 0.783 & +3.6 & +5.0 & \textcolor{tabmuted}{\textminus{}0.3} & +6.7 & +11.0 & \textcolor{tabmuted}{+0.9} & +3.1 & 0.767 \\
Perplexity Decider-27B & 0.798 & +6.2 & +3.3 & +2.6 & +10.7 & +10.4 & \textcolor{tabmuted}{+2.0} & +4.8 & 0.873 \\
Torchcast-27B & 0.486 & +21.3 & +15.5 & +16.3 & +27.5 & +34.6 & +14.1 & +6.4 & 0.601 \\
SimpleJev-27B & 0.474 & +19.3 & +3.0 & +4.5 & +12.0 & +18.8 & \textcolor{tabmuted}{+1.2} & +7.0 & 0.638 \\
\midrule
\multicolumn{10}{@{}l}{\emph{System One readout of a frozen LLM}} \\
Decider-chat-31B & 0.762 & +8.2 & \textcolor{tabmuted}{+2.3} & +4.2 & +10.4 & +10.2 & \textcolor{tabmuted}{+2.0} & +5.9 & 0.871 \\
\midrule
\multicolumn{10}{@{}l}{\emph{LLM, likelihood readout (thinking off)}} \\
Gemma-4-31B & 0.775 & +7.0 & \textcolor{tabmuted}{\textminus{}0.6} & \textcolor{tabmuted}{+1.7} & +6.4 & +8.7 & \textcolor{tabmuted}{\textminus{}0.9} & \textcolor{tabmuted}{\textminus{}1.2} & 0.795 \\
Qwen3.8-27B & 0.783 & +9.9 & +5.7 & +3.7 & +6.1 & +9.6 & +4.8 & +6.2 & 0.360 \\
\midrule
\multicolumn{10}{@{}l}{\emph{LLM, reasoning (reference)}} \\
GPT-6.1 & 0.953 & +1.4 & \textcolor{tabmuted}{+0.9} & +2.5 & \textcolor{tabmuted}{+0.9} & +1.9 & +1.6 & \textcolor{tabmuted}{+1.1} & 0.352 \\
\bottomrule
\end{tabular}
\caption{Decision-oriented blocks on 644 ALFWorld states: accuracy alone, change (points) with each block (correct values unless marked own) and accuracy on the sub-goal question. Gray: 95\% interval includes zero. Quyet-1.0-Large is omitted (answers 2 of 644).}
\label{tab:dprobes}
\end{table*}

%% file: sections/08_controls.tex
\section{Controls, Boundaries and Calibration}
\label{sec:controls-results}

The paired outcomes describe single states. The controlled variations of \S\ref{sec:controls} ask what
the models rely on, where their accuracy begins to fall, and whether their distributions can be used to
decide when to trust them.

\begin{figure*}[!tp]
\centering
\includegraphics[width=\textwidth]{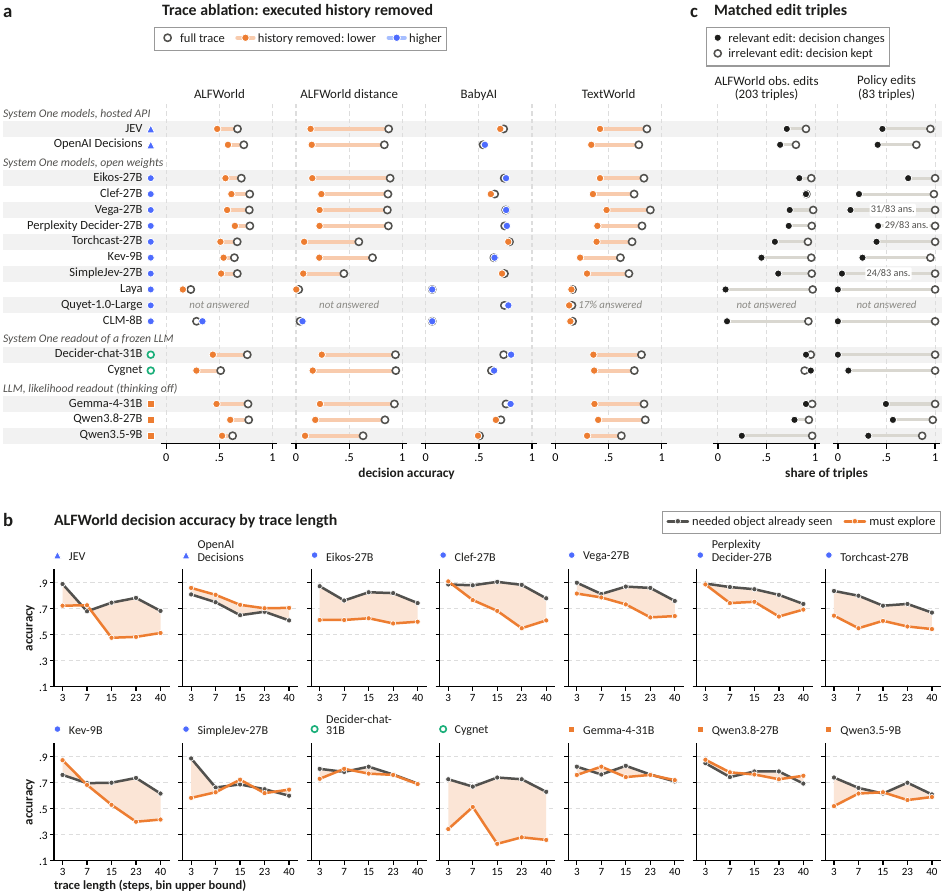}
\caption{Controlled variations. (a) Accuracy with the full trace (ring) and with the history removed (dot). (b) ALFWorld
accuracy by trace length, for states whose needed object was already seen and states that require exploration.
(c) Matched edit triples: share of relevant edits that change the decision (filled) and of irrelevant edits that
keep it (ring). Systems failing the coverage rule are marked or omitted.}
\label{fig:controls}
\end{figure*}

\paragraph{The history is used where it is needed.} Removing the executed history leaves BabyAI almost
unchanged (JEV 73.3$\to$70.1\%; Gemma even rises from 75.5\% to 79.9\%), since the current view largely
determines the next step, but it collapses the states that require remembering an earlier observation
(Figure~\ref{fig:controls}a). On the ALFWorld distance ladder, every strong system falls to between 14\%
and 24\% (JEV 87.2$\to$13.7\%, Gemma 92.6$\to$22.6\%, Clef 86.5$\to$23.9\%, Cygnet 93.9$\to$15.7\%), and on
TextWorld to between 35\% and 42\% (from 74--86\%). The other systems that answer ALFWorld decisions
fall into the same band on the distance ladder (OpenAI Decisions 83.1$\to$14.8\%, Eikos-27B
88.6$\to$15.3\%, Perplexity Decider-27B 86.9$\to$22.1\%, Vega-27B 86.0$\to$22.2\%, Decider-chat-31B
93.6$\to$24.3\%; SimpleJev-27B and Torchcast-27B, weaker to begin with, 45.1$\to$6.8\% and 59.1$\to$7.7\%) and to 30--48\% on
TextWorld, while on BabyAI they lose at most 2.3 points or improve. The models therefore do read the trace, and the
facts-right, action-wrong errors of \S\ref{sec:results} are not a sign that the history is ignored.

\paragraph{Exploration degrades with trace length; distance does not.} The ALFWorld ladders locate where
accuracy begins to fall (Figure~\ref{fig:controls}b). When the needed object has already been seen,
Clef and Gemma stay within about ten points of their short-trace accuracy up to 23 steps, while JEV falls
from 0.89 for three or fewer steps to 0.68--0.78 beyond. The larger effect is in states that still require
exploration: JEV drops from 0.72 to 0.48 once the trace exceeds seven steps and Clef from 0.91 to 0.55,
while Gemma stays between 0.72 and 0.82. Keeping track of which places have already been searched is the
hard part. The distance between the decisive observation and the decision, in contrast, has no effect over
the range we test: from zero to 15 irrelevant intervening steps, no competent system loses more than three
points (e.g.\ Gemma 0.92--0.94, JEV 0.84--0.90). The other systems agree: all lose accuracy on long traces
(OpenAI Decisions even when the object has been seen, 0.81 to 0.61), and none loses accuracy with distance;
Torchcast-27B even gains (0.52 with no intervening step, 0.64 with 15).

\paragraph{Relevant edits are noticed in simulators, missed in policies.} When a single observation is
edited so that the gold action changes (Figure~\ref{fig:controls}c), Cygnet, Clef and Gemma change their
decision in 91--96\% of ALFWorld triples, Qwen3.8-27B in 79\% and JEV in 71\%, while all keep
their decision after an irrelevant edit in 89--97\% of cases. Decider-chat-31B
(91\%) and Eikos-27B (84\%) react about as often as Cygnet, Clef and Gemma, and OpenAI Decisions is the least
stable, changing after 64\% of relevant edits but also after 20\% of irrelevant ones; Torchcast-27B changes its
decision after only 59\% of relevant edits. Customer-service policy edits are a sharper
case of the dissociation. When one record field is edited so that the policy now requires a different tool
call, JEV, Clef, Gemma and Qwen3.8-27B answer every probe on the edited state correctly (all 374 edited
states), yet in the 83 matched triples they change the call after a relevant edit in only 22\% (Clef) to
57\% (Qwen3.8-27B) of cases. The models read the edited fact and still act as if it
had not changed. OpenAI Decisions answers 99.5\% of the edited states' probes correctly and changes the call
in 41\%; Torchcast-27B answers all of them correctly and changes the call in 40\%. Eikos-27B is the exception: with every probe right, it changes the call after 72\% of relevant
edits and never after an irrelevant one. Decider-chat-31B never selects the required tool call on these
states, so the question does not arise for it.

\paragraph{Stated rule changes are followed.} When a TextWorld game states that cooking rules differ from
the usual ones, the strong systems follow the stated rules in 90--99\% of states (Clef 99.3\%, Qwen3.8-27B
96.3\%, JEV 93.3\%, Gemma 90.3\%), with the same results on a cohort generated later from new seeds. Kev-9B
reaches 83.0\% on the same states without ever answering all the rule probes correctly, an example of a
correct action that does not follow from the measured facts. The other System One models follow the stated rules in
89--100\% of states in both cohorts, with every rule probe answered correctly.

\begin{figure*}[!tp]
\centering
\includegraphics[width=\textwidth]{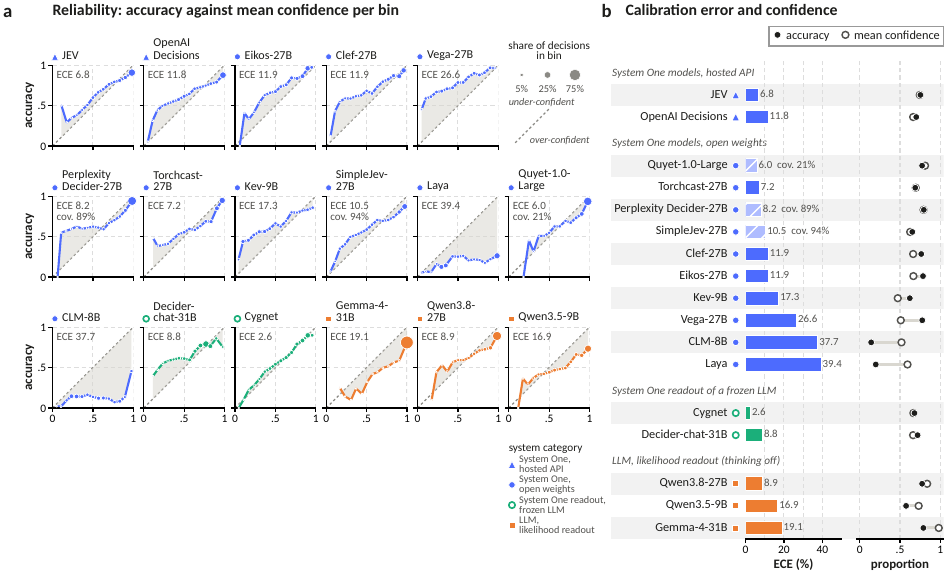}
\caption{Calibration of native decision distributions on test decisions. (a) Reliability diagrams; marker area is the
share of decisions in the bin. (b) Expected calibration error (15 bins; hatched: coverage below 95\%) and accuracy
against mean confidence. GPT-6.1 is excluded (verbalized probabilities).}
\label{fig:calibration}
\end{figure*}

\paragraph{Calibration differs more than accuracy.} Because System One models return a distribution, their
confidence could decide when to escalate. Figure~\ref{fig:calibration} shows that calibration varies far
more across systems than accuracy does. Cygnet, whose recipe includes a calibration step, has an expected
calibration error of 2.6\%, and JEV 6.8\%. Clef and Kev-9B are underconfident (mean confidence 0.66 and 0.47
at accuracies of 0.76 and 0.62), while the likelihood readout of Gemma is almost always near-certain (mean
confidence 0.98 at an accuracy of 0.79; ECE 19.1\%). For JEV, confidence is informative enough to drive the
cascade of \S\ref{sec:ladder}, but for an uncalibrated readout the same threshold would rarely fire.
OpenAI Decisions, Eikos-27B and Decider-chat-31B are underconfident like Clef (mean
confidence about 0.66 at accuracies of 0.70--0.79; ECE 8.8--11.9\%), and Perplexity Decider-27B is close
to Qwen3.8-27B (8.2\%). Torchcast-27B is about as well calibrated as JEV (ECE 7.2\%; mean confidence 0.70 at
an accuracy of 0.69). OpenAI Decisions' own confidence field, which we do not use, is lower still (ECE
13.9\%). Quyet-1.0-Large's low error (6.0\%) covers only the 21\% of decisions it answers. Apart from CLM-8B and
Laya, Vega-27B is the least calibrated system we measure (ECE 26.6\%): its mean confidence of 0.51 at an accuracy of 0.78 makes it
strongly underconfident.

%% file: sections/02_related.tex
\section{Related Work}
\label{sec:related}

\paragraph{System One models and next-action evaluation.} Scoring a supplied set of answers, long used in
classification and reranking, is now offered as a programmable interface for agents, by hosted models
\citep{typesafe2026jev,openai2026decisions,liquid2026d1} and open alternatives
\citep{clm2026,kev2026,clef2026,laya2026,eikos2026,pplxdecider2026,simplejev2026,vega2026,quyet2026,cygnet2026,deciderchat2026}.
As an evaluation judge, JEV approaches a reasoning judge where a verdict can be read off the text but falls behind
where it must be derived, and its confidence can decide when to escalate \citep{li2026jevjudge}. Public comparisons report aggregate scores \citep{decisionindex2026}; agent benchmarks measure task success, progress
or per-step correctness \citep{shridhar2021alfworld,ma2024agentboard,barres2025tau2bench,deng2023mind2web}, and fixed
histories with a next-action question are established \citep{oedd2024}. We ask what such accuracy leaves
unexplained, pairing each decision with probes on the same state.

\paragraph{Knowing versus doing, and composition.} Studies of agents
\citep{schmied2025greedy,sobotka2026,stocktake2026,goaldirected2026} separate what an agent appears to know from what
it does. Language models are also distracted by irrelevant context \citep{shi2023distracted}, plan better with an
external component \citep{valmeekam2023planning,kambhampati2024llmmodulo}, and answer every sub-question of a problem
yet fail the composed question \citep{press2023compositionality}. Our setting exposes neither text nor hidden states,
the probes are answered in isolation on the exact state, and we test whether the probe set can support the diagnosis
at all. The remedies we test mirror decomposed and program-aided reasoning
\citep{zhou2023leasttomost,khot2023decomposed,gao2023pal}, but the sub-questions go to the same choice-only model and
the composition is done by the calling program.

\paragraph{Behavioral diagnosis.} Chains of thought need not be faithful to the computation behind an answer
\citep{turpin2023unfaithful,lanham2023faithfulness}. Like behavioral testing \citep{ribeiro2020checklist}, and unlike
probing classifiers \citep{belinkov2022probing}, we infer competence from controlled questions and input variations.
Appendix~\ref{app:related} discusses each line of work, cost-aware routing and failure attribution in full.

%% file: sections/09_discussion.tex
\section{Discussion}
\label{sec:discussion}

\paragraph{What a practitioner can do.} The results suggest a simple procedure for deploying a System
One model in an agent loop. First, measure the paired outcomes on the target states. If most errors
follow misread facts, the remedy lies in the state representation, the trace format or the context budget.
If most follow correct reading, as for every strong system here, the next question is whether the
decision-relevant facts are known and the rules explicit. Where they are, as on our generated procedures,
the strongest remedy is to ask the facts as narrow parallel questions and compose the action in code,
which brought every system we tried to 0.96--1.00 at the price of a second batch of cheap questions; a
confidence cascade keeps most of the gain with a second request on a third of the states. Where the
decision requires choosing a sub-goal in an open task, as in ALFWorld, appending facts helps little, and the
missing ingredient is a sub-goal that the model does not reliably supply; that is a planning problem, for
a planner, for training, or for a deliberating model on the few states that need it.

\paragraph{A diagnostic instrument, not a leaderboard.} The pooled accuracies of
Table~\ref{tab:paired} hide most of what matters. Rankings change across domains
(Figure~\ref{fig:domains}); a perfect action score on coding traces hides a twelve-point difference in
reading; two systems with similar paired accuracy fail for different reasons (consolidation for JEV,
cue capture for Clef); and the most accurate fast system, Gemma's likelihood readout, is among the least
calibrated. Fast decision models are therefore better evaluated with diagnostic panels and their controls, and
with freshly generated states whose probes are sufficient by construction, that match the kind of state a workflow
depends on, than with a single number.

\paragraph{Fast is not the same as cheap to trust.} GPT-6.1 shows what deliberation buys on the same
states: seven to twelve points of decision accuracy at roughly seventy times the cost of JEV. It does
not change the kind of error that remains. For agent builders, the interesting design space is therefore
not System One versus reasoning models, but which states a fast model can be trusted on, which need a
composed decision, and which need deliberation. The paired measurement, the probe-sufficiency check and the
model's own calibration are the instruments for drawing that line.

%% file: sections/10_closing.tex
\section{Limitations}
\label{sec:limitations}

\emph{Behavioral evidence only.} The paired outcomes are observations on identical inputs. They do not
show which internal computation produced an action, and a correct probe answer does not prove that the
model used that fact. \emph{Probe coverage.} On natural states the probes are not sufficient for the
decision, so a facts-right, action-wrong outcome mixes decision failures with facts we did not ask about;
the generated ladder and the decision-oriented probes address this only for the domains they cover.
\emph{Labels.} Gold actions are computed from agent-observable information, but they can be stricter than
necessary (5--42\% of a system's simulator errors are in fact optimal) and, in the original ALFWorld cohorts,
co-located receptacles produce wrong location labels for 303 of 8,197 paired ALFWorld records; we keep the frozen
labels and report sensitivity analyses. \emph{Deployments, not architectures.} Systems differ in serving
stack, readout and input limits, two author runtimes shorten long states themselves, and hosted models can
change between versions (96.3\% agreement on a JEV rerun after five days; OpenAI Decisions is a public beta). \emph{Scope.} We evaluate single decisions on frozen traces, not closed-loop
execution, and the follow-up studies of \S\ref{sec:mechanism} cover a subset of systems: the composition
ladder and probe-then-decide sixteen each, and the decision-oriented probes twelve (each count includes the
reasoning reference). The reasoning reference answered each
question in a separate call with a verbalized distribution, so its calibration is not comparable.

\section{Conclusion}
\label{sec:conclusion}

A fast decision model that errs leaves nothing behind but a wrong action. By asking the same model, on the same
state, what it read and what it chose, we find that for strong systems the problem is usually not reading. Across
seventeen fast systems and a reasoning model, most wrong actions follow correct reading, and they gather on states
that defeat systems of every kind. Reading right and acting wrong is therefore not the mark of weak models but the
characteristic failure of strong ones. Nor is it a single failure. When the decisive facts are known, models fail
to combine them, and their own answers, composed by code, repair the decision; in open tasks they lack the next
sub-goal, which additional facts do not supply. Both point to one way of building agents on fast models:
let the model read, through many narrow questions, and hand the composition of the decision to code, to a
planner, or to deliberation where a state requires it. Measuring reading and acting separately is what tells
which a state requires.

%% file: sections/appendix_related.tex
\section{Extended Related Work}
\label{app:related}

\paragraph{System One models for agents.}
Most language agents generate reasoning and actions token by token
\citep{yao2023react,wei2022cot}. System One models instead score a supplied
set of answers and return a distribution over them, a pattern long used in
classification and reranking that is now offered as a general programmable
interface for agents and software. Examples include hosted models such as JEV \citep{typesafe2026jev}, OpenAI's
Decisions API \citep{openai2026decisions} and Liquid's decision models
\citep{liquid2026d1}, and open alternatives that range from a two-tower head over
frozen embeddings \citep{clm2026} to fine-tuned pointer and schema heads
\citep{kev2026,clef2026}, a compact encoder \citep{laya2026}, fine-tuned 27--31B
decision models \citep{eikos2026,pplxdecider2026,simplejev2026,vega2026,quyet2026},
and recipes that read a frozen open-weight model through the same interface
\citep{cygnet2026,deciderchat2026}. Public comparisons of these systems report aggregate
scores on decision suites \citep{decisionindex2026}. We instead ask what
such accuracy leaves unexplained: whether errors reflect misreading or
mis-deciding, and how far that question can be answered through a choice-only
interface. The ``System~1'' label borrows the fast/deliberate distinction of
\citet{kahneman2011thinking}; we use it, as these models' developers do, only
to name the interface, not to claim anything about the computation inside the
models. Their output types go beyond choices; we evaluate only the choice
interface.

\paragraph{Agent benchmarks and next-action evaluation.}
Interactive environments such as ALFWorld \citep{shridhar2021alfworld},
BabyAI \citep{chevalierboisvert2019babyai,paglieri2025balrog} and TextWorld
\citep{cote2018textworld,tales2025}, and suites such as AgentBoard
\citep{ma2024agentboard}, $\tau$-bench \citep{yao2024taubench,barres2025tau2bench}
and Mind2Web \citep{deng2023mind2web}, mainly measure task success, progress or
per-step action correctness. Tool-use benchmarks test tool selection, argument
filling and when to call a tool \citep{patil2025bfcl,shi2025toolret,ross2025when2call},
and large synthesized tool-use trajectories are available for training \citep{toucan2025},
and trajectory-judgment benchmarks test whether a model can tell success,
safety or the failing step from a finished trace
\citep{lu2025agentrewardbench,yuan2024rjudge,zhang2025whowhen}. Presenting a
fixed agent history and asking for the next action is itself established
\citep{oedd2024}; AgentRM-style trees provide sibling actions with value
estimates \citep{xia2025agentrm}. We reuse many of these sources, but we pair
each decision with independently answered factual probes on the same state,
place every admissible action in context, and compute gold labels only from
information visible to the agent. Label errors in test sets can change
benchmark conclusions \citep{northcutt2021pervasive}; re-deriving labels from
the agent-visible text, as we do in \S\ref{sec:labels}, is one way to find
them when a simulator's internal state and the text disagree.

\paragraph{Knowing versus doing.}
Several recent studies separate what an agent appears to know from what it
does. \citet{sobotka2026} distinguish observation--belief from belief--action
gaps and intervene on beliefs through steering and prompting.
\citet{stocktake2026} measure state judgments and control separately in a
supply-chain task, \citet{goaldirected2026} combine behavior, optimal policies
and representation probes in grid worlds, and \citet{schmied2025greedy} report
a knowing--doing gap in which models can state a good strategy but do not act
on it. State-maintenance methods for long-horizon agents decide when to
compress, retrieve or rewrite history \citep{pamer2026,statecomp2026}. Our
setting differs in three ways that follow from the interface we study. The
models expose neither text nor hidden states, so all evidence is behavioral;
the probes are answered in isolation on the exact decision state rather than
elicited as free-text rationales; and we test whether the probe set is
sufficient to support the diagnosis at all. At the scale of a full benchmark,
we find the knowing--doing gap to be the dominant source of error for every
strong system, including a reasoning model, and we show that it is not one
phenomenon: depending on system and state, it reflects a failure to combine
facts, capture by irrelevant cues, or the choice of the next sub-goal.

\paragraph{Composing what has been read.}
Language models often answer every sub-question of a multi-step problem yet
fail the composed question, a \emph{compositionality gap} that does not shrink
with model size, because single-step accuracy improves faster than composed
accuracy \citep{press2023compositionality}.
They are also distracted by irrelevant context \citep{shi2023distracted} and use
evidence unevenly across long inputs \citep{liu2024lost}. Our composition ladder
(\S\ref{sec:ladder}) is a controlled version of this question for agent
decisions: with probes that provably determine the action, we can attribute a
facts-right, action-wrong outcome to composition, and the rewrites separate
consolidation of scattered evidence from capture by irrelevant cues. The
remedies we test mirror known ones. Answering sub-questions first and then
deciding resembles least-to-most and decomposed prompting
\citep{zhou2023leasttomost,khot2023decomposed}, and composing the action in code
resembles program-aided reasoning \citep{gao2023pal}. The difference is the
setting: the sub-questions are asked of the same choice-only model in parallel,
without generated text, and the composition is done by the calling program, as
System One models are designed to be used \citep{typesafe2026jev}.

\paragraph{Planning with language models.}
Language models are weak autonomous planners, and planning works better when a
model proposes and an external component checks or plans
\citep{valmeekam2023planning,kambhampati2024llmmodulo}. On real ALFWorld states
we find a matching pattern for a System One model: JEV reads the state almost
perfectly, but stating the next sub-goal adds 16 points that no amount of
additional facts provides (\S\ref{sec:dprobes}). This suggests a division of
labor in which the fast model reads and acts, and the sub-goal comes from a
planner or a deliberating model.

\paragraph{Cost-aware routing.}
Cascades that send only uncertain inputs to a more expensive model reduce cost
at little loss in accuracy \citep{chen2024frugalgpt}, provided the score that
decides whether to accept the cheap answer is informative. The same idea has been applied to evaluation with a
System One model: JEV used as a judge accepts its confident verdicts and escalates the rest to a reasoning judge
\citep{li2026jevjudge}. Our confidence cascade (\S\ref{sec:ladder}) applies
this idea within one System One model, and our calibration results
(\S\ref{sec:controls-results}) show that whether it can work depends strongly on
the readout.

\paragraph{Failure attribution and explanation faithfulness.}
Automated failure attribution asks which agent or step caused the failure of a
single- or multi-agent system \citep{zhang2025whowhen,whowhenpro2026}, and recent work stresses that
an error can admit several valid explanations \citep{mpbench2026}. Our target
is narrower and lower in the stack: for a single decision, was the available
evidence read correctly? Generated explanations are an unreliable route to
this answer, because chains of thought need not be faithful to the
computation that produced an answer
\citep{turpin2023unfaithful,lanham2023faithfulness}. Our method is closer to
behavioral testing \citep{ribeiro2020checklist}: we infer competence from
controlled questions and input variations, and, unlike probing classifiers
\citep{belinkov2022probing}, we need no access to internal representations.
Finally, because System One models expose probabilities, we also ask whether
their confidence flags risky decisions \citep{guo2017calibration,kadavath2022know}.

%% file: sections/appendix_measurement.tex
\section{Measurement Details}
\label{app:setting}

\subsection{Decision states}
\label{app:states}

A decision state $s$ consists of a task goal $g$, the executed history
$h_{1:t}=(a_1,o_1,\dots,a_t,o_t)$ of actions and the observations they
produced, any rules $R$ that govern the environment or policy, and the set
$\mathcal{A}(s)$ of admissible next actions. The model receives $s$ and must
choose one element of $\mathcal{A}(s)$; it is never asked to generate an
action. We call this the \emph{closed-world} principle: every admissible
action is placed in context with its canonical name and a description, nothing
is shortlisted, and the gold action is never the only well-formed option.
Actions with open arguments, such as tool calls, are enumerated from the
environment or database. When a flat list would exceed the interface limit of
255 options, the call is factorized into one choice for the tool and one per
closed-set argument, and scored as a complete call. Candidate sets come from
the environment, from code, or from the dataset as released; no language model
proposes candidates at evaluation time, since that would measure the proposer
rather than the System One model.

Gold is a \emph{set} $G(s)\subseteq\mathcal{A}(s)$ of acceptable actions,
computed only from information available to the agent. In the simulated
environments, $G(s)$ contains the first actions of all shortest plans on a map
built from what the trace entails. States where the needed objects have not
yet been observed are labeled as exploration states, whose gold is the set of
actions that reveal unseen locations (\S\ref{sec:benchmark}). A decision is
correct, $D(s)=1$, if the chosen action lies in $G(s)$.

\subsection{Factual probes}
\label{app:probes}

On annotated states we additionally ask factual probes $q_1,\dots,q_m$ about
the same state. Their answers are determined by the simulator state, the
database or the data labels:

\begin{center}
\small
\begin{tabular}{@{}L{0.3\columnwidth}L{0.62\columnwidth}@{}}
\toprule
Probe family & Example \\
\midrule
Feedback outcome & Did the last action succeed, fail, or have no effect? \\
Current state & Where is the agent? What is it holding? \\
Progress & Which subgoals are complete? \\
Constraint & Is action $x$ permitted in the current state? \\
Evidence & Which earlier observation establishes this fact? \\
\bottomrule
\end{tabular}
\end{center}

Which families apply depends on the domain, and we report coverage per domain
(\S\ref{sec:benchmark}). Probes are closed questions with all candidate
answers listed. Each question, whether probe or decision, is answered in
isolation. Several questions about one state may share a request, as the
interface allows, but each is answered on the state alone: no question or
answer is added to another question's input, and no gold fact is revealed. ``Paired'' therefore means that the input state is
identical. It does not mean that we observe successive stages of the model's
computation.

\subsection{Four observable outcomes}
\label{app:outcomes}

For a state $s_i$ with applicable probes, let $\U_i=1$ if all of them are
answered correctly, and let $\D_i$ be decision correctness. $\U_i$ is
\emph{measured-fact performance}, not a latent variable of complete
understanding. We also record the mean probe accuracy of each state, and we
write U1D0 for the outcome $\U{=}1,\D{=}0$ (and U1D1, U0D1, U0D0 likewise). Crossing
$\U_i$ and $\D_i$ gives four outcomes (Table~\ref{tab:outcomes}).

\begin{table}[t]
\centering
\small
\begin{tabular}{@{}ccL{0.33\columnwidth}L{0.33\columnwidth}@{}}
\toprule
$\U$ & $\D$ & Directly observed & Not licensed \\
\midrule
1 & 1 & Measured facts and action both correct & That all relevant facts were understood \\
0 & 0 & Some probe wrong; action wrong & That the probe error caused the wrong action \\
1 & 0 & All measured facts correct; action wrong & That the failure lies purely in deciding \\
0 & 1 & Some probe wrong; action correct & That the model used a shortcut, or that the wrong fact mattered \\
\bottomrule
\end{tabular}
\caption{The four observable outcomes on a paired state and the conclusions
each does and does not support.}
\label{tab:outcomes}
\end{table}

We report the four joint rates together with $P(\D{=}1\mid \U{=}1)$ and
$P(\D{=}1\mid \U{=}0)$, mean probe accuracy, and the number of applicable
states. Because different models reach $\U{=}1$ on different states,
conditional rates also reflect composition. We therefore read them alongside
the joint table on identical states, and leave a conditioning event that is
empty undefined rather than reporting zero. Questions a system cannot answer
because of an input limit stay in the denominator as wrong, and are reported
separately as coverage, since a deployment limit is not a cognitive error.

\paragraph{Estimation.}
Records come in difficulty cells nested in task groups. The main text reports
raw counts pooled over all 19,663 paired test states (or over the states of a
study). These counts mix heterogeneous tasks and probe granularities, so they
summarize prevalence and are not a ranking statistic. Cell-weighted estimates
instead average probes within a record, records
within a frozen cell, and cells equally within a panel, with percentile
intervals from 2,000 paired bootstrap draws that resample whole source
clusters (games, users, repositories), since states from the same source are
dependent. Intervals shown for the follow-up studies use the same idea with
the study's own clusters (\S\ref{sec:mechanism}).

\subsection{When can a probe set support the diagnosis?}
\label{app:sufficiency}

An outcome with $\U{=}1,\D{=}0$ suggests a decision failure only if the probes
capture what the decision requires. This can be checked without looking at the
model at all. Hold fixed the goal, the rules and the remaining context, and let
$P(s)$ be the vector of correct probe answers for state $s$ and $G(s)$ its set
of acceptable actions. For a group of states $S$ that share a probe vector,
the probe set is \emph{sufficient on} $S$ if
\begin{equation}
\bigcap_{s\in S} G(s)\neq\emptyset,
\label{app:eq-sufficiency}
\end{equation}
that is, if some action is correct no matter which of these states the agent
is in. If two states have the same correct probe answers but disjoint gold
sets, then answering every probe correctly cannot by itself determine the
right action, so a wrong action after correct probes cannot be attributed to
deciding alone. Pairwise intersections are not enough; the check must
consider the whole group. In finite, fully enumerable environments it can be
run exhaustively. In natural data, finding no counterexample in a sample does
not prove sufficiency. Context that the probes do not ask about, such as rules
or tool availability, must either be held fixed or be probed. Passing the
check means the probes \emph{can} support a diagnosis. It does not show that
the model actually uses those facts.

\subsection{Controlled variations}
\label{app:controls}

The paired outcomes describe a single state. To see what drives them, the
benchmark also contains controlled variations of states:
\begin{itemize}
\item a \emph{trace ablation} that removes the executed history and keeps the
goal and current observation;
\item observation \emph{minimal pairs} that edit only the last observation,
either relevantly, which changes the gold action, or irrelevantly, which does
not;
\item \emph{rule changes} stated in the environment description, such as
altered recipes or verbs, that make prior knowledge wrong;
\item \emph{encoding controls}: option-order permutations, named versus opaque
action identifiers with descriptions kept, flat versus factorized tool calls,
and conversation-history rendering;
\item \emph{difficulty ladders} over trace length, distance from the decisive
observation to the decision, size of the action set, and novelty, from which
we read capability boundaries.
\end{itemize}
This paper reports the trace ablation, the minimal pairs, the rule changes and
the trace-length and distance ladders (\S\ref{sec:controls-results}); the
encoding controls and the remaining ladders are not analyzed here.

%% file: sections/appendix_benchmark.tex
\section{Benchmark Construction}
\label{app:benchmark}

\input{tables/composition}

Table~\ref{tab:composition} summarizes the frozen benchmark. It has 65,228
records: 59,234 test and 5,994 dev. Agent decision states are the core; a
supporting track of text-understanding primitives measures abilities that
agent decisions draw on, such as multi-hop and long-context reading. We
describe the agent states here.

\subsection{Simulated decision states}
\label{app:sim}

Simulators give exact ground truth for both actions and facts, so most paired
states come from three of them. In ALFWorld \citep{shridhar2021alfworld} we
use the unseen evaluation games, plus the seen games as a separate cell; in
BabyAI-Text we use BALROG's text rendering \citep{paglieri2025balrog,chevalierboisvert2019babyai,carta2023glam};
and in TextWorld we generate cooking games at four TALES difficulty settings
\citep{cote2018textworld,tales2025}. Trajectories are rolled out by the
environment's own expert, by a noisy variant that takes a uniformly random
admissible action with probability 0.2, and by a variant that issues invalid
commands. This puts failures, error feedback and detours into the traces,
which successful demonstrations rarely contain. No language model takes part
in the roll-outs.

Gold actions use only what the trace reveals. For each state we build a map
from the observations so far and compute the first actions of all shortest
plans. A state is \emph{determinable} when the needed objects have already
been observed and the gold set is the same whether unseen cells are assumed
free, assumed blocked, or taken from the true map. Otherwise it is an
exploration state, whose gold is the set of actions that reveal unseen
locations. Sampled items were checked against breadth-first search in the
real environments; for example, 234 of 234 BabyAI items and 63 of 63 TextWorld
items matched. Probe answers are read from the simulator state.

On these environments we build the ladders and controls of
\S\ref{sec:controls}. ALFWorld cells vary trace length, and a distance ladder
inserts valid but irrelevant steps between the decisive observation and the
decision. Observation minimal pairs move a sighting to another receptacle.
Novelty cells rename objects and verbs or replace action names with opaque
identifiers while keeping their descriptions. BabyAI and TextWorld cells
follow the environments' difficulty levels. Confirmation cohorts generated
later from new seeds re-test the main findings on states that no earlier run
had seen. Separate TextWorld cohorts change the rules of cooking in the
game description, so that prior knowledge about recipes no longer applies.

\subsection{Policy, tools, web and coding}
\label{app:offline}

\paragraph{Customer service.} From $\tau^2$-bench airline and retail
\citep{barres2025tau2bench} we take decision states from recorded and native
trajectories. Tool calls are enumerated from the task database, and probes ask
about record fields and whether a policy permits an action. Minimal pairs edit
one policy-relevant field so that the correct call changes, or edit an
irrelevant field so that it does not. A native recovery panel places the agent
after a failed or incomplete call.

\paragraph{Offline traces and tools.} AgentRM trees supply all sibling actions
at a node together with value estimates \citep{xia2025agentrm}. Mind2Web
supplies all candidate page elements and operations \citep{deng2023mind2web}.
Toucan trajectories, ToolRet, BFCL and When2Call supply tool selection over full tool lists
that include tools the model has not seen, as well as the choice between
calling a tool, asking the user, and abstaining
\citep{toucan2025,shi2025toolret,patil2025bfcl,ross2025when2call}. These
sources have no native comprehension labels. A separate panel adds
\emph{visible-field} probes, whose answers are stated explicitly in the trace,
to a subset of them.

\paragraph{Coding traces.} To test feedback reading in software agents
without asking the model to write code, we build 360 records from real
SWE-smith repository images. Each case combines two genuine patches to
different files in the same repository, runs the published tests, and asks
which edit caused a named check to regress, whether the displayed evidence is
current or stale, and which limited next action follows. The 50 test source
pairs span 17 repositories, which are disjoint from the 11 development
repositories.

\paragraph{Trace judgments.} AgentRewardBench, R-Judge and Who\&When ask for
judgments about a completed trajectory, namely success, side effects, safety
and the failing step \citep{lu2025agentrewardbench,yuan2024rjudge,zhang2025whowhen}.
We report them as a separate family, since they are not next-action
decisions.

\subsection{Text-understanding primitives}
\label{app:trackb}

Twenty-four task groups (32,700 test records) measure negation, multi-hop
reasoning, long context with distractors, multi-turn recollection and
correction, new label definitions, counterfactual rules, and logical
reasoning. Each group is arranged along a difficulty ladder, for example hop
count or context length. Sources include CondaQA, ScoNe, ProofWriter, CLUTRR,
MuSiQue, BABILong, CLINC150, MASSIVE, CLadder and BoardgameQA, together with
programmatic generators
\citep{ravichander2022condaqa,she2023scone,tafjord2021proofwriter,sinha2019clutrr,trivedi2022musique,kuratov2024babilong,larson2019clinc,fitzgerald2023massive,jin2023cladder,kazemi2023boardgameqa}.
These records have no paired probes. They allow boundaries in general reading
to be compared with boundaries in agent decisions; this paper
does not analyze them and does not use them as a measure of next-action skill.

\subsection{Splits, freezing and audits}
\label{app:freeze}

Records are split into dev and test by a hash of their source group (game,
user, repository), so no source contributes to both splits. Dev is used for
template and pipeline checks and for selecting the reasoning reference, and is
always reported separately from test. The target is 300 test records per
difficulty cell, sampled with matched pairs across cells where a ladder
requires them. The full population, including every request byte, option
order and label, is frozen in a hash-sealed manifest before analysis, and all
systems receive the identical requests. An independent content audit found 42
test records with conflicting labels outside the control groups and 56 test
records whose inputs also occur in dev. We keep the frozen population
unchanged and report a sensitivity analysis that excludes these 98 records.
Public training games cannot be assumed unseen by pretrained models, which is
one reason we add freshly generated confirmation cohorts. Fixed public items
can also be overfit. The generators we use for the follow-up studies, including one for decision states
whose probes are sufficient by construction (\S\ref{sec:ladder}), can draw new evaluation states at any time.

%% file: tables/composition.tex
\begin{table*}[!tp]
\centering
\small
\setlength{\tabcolsep}{4.5pt}
\renewcommand{\arraystretch}{1.12}
\begin{tabular}{@{}L{2.45cm}L{4.0cm}rrrL{5.3cm}@{}}
\toprule
 & & \multicolumn{3}{c}{Records} & \\
\cmidrule(lr){3-5}
Domain & Sources & Test & \begin{tabular}[b]{@{}r@{}}Paired\\test\end{tabular} & Dev & Decision / probes \\
\midrule
\multicolumn{6}{@{}l}{\emph{Agent decision states}} \\
Household (sim.) & ALFWorld unseen and seen games & 9,097 & 9,097 & 840 & admissible commands; location, holding, last outcome, progress \\
Grid world (sim.) & BabyAI-Text (BALROG) & 2,181 & 2,181 & 236 & 6 actions; orientation, holding, target position, last outcome \\
Text game (sim.) & TextWorld cooking (TALES levels) & 2,400 & 2,400 & 300 & admissible commands; room, holding, ingredient state, item location \\
Customer service & $\tau^2$-bench airline, retail & 3,739 & 3,739 & 330 & enumerated tool calls; record fields, policy constraints \\
Coding & SWE-smith repository traces & 300 & 300 & 60 & next action; edit effect, current vs.\ stale evidence, attribution \\
Workflow & rule-governed synthetic workflows & 240 & 240 & 40 & next call; state and constraint facts \\
Offline traces & AgentRM, Mind2Web & 1,800 & -- & 180 & sibling actions; page elements and operation \\
Tools & Toucan, ToolRet, BFCL, When2Call & 2,400 & -- & 240 & full tool list; call vs.\ ask vs.\ abstain \\
Visible-field probes & parents in AgentRM, Mind2Web, tools & 726 & 726 & 59 & as parent; facts stated in the trace \\
Encoding controls & permuted, rendered, factorized copies & 2,570 & 980 & 320 & as parent \\
Trace judgments & AgentRewardBench, R-Judge, Who\&When & 1,081 & -- & 119 & success, side effects, safety, failing step \\
\midrule
\multicolumn{6}{@{}l}{\emph{Text-understanding primitives}} \\
24 task groups & negation, multi-hop, long context, multi-turn, label definitions, rules, reasoning & 32,700 & -- & 3,270 & native task labels \\
\midrule
Total & & 59,234 & 19,663 & 5,994 & \\
\bottomrule
\end{tabular}
\caption{Composition of the frozen benchmark (65,228 records). ``Paired test''
counts test states that carry factual probes. Dev records are reported
separately and never pooled with test results.
Household includes the targeted-coverage cohort; grid-world and text-game rows
include newly generated confirmation cohorts and, for TextWorld, native rule
changes.}
\label{tab:composition}
\end{table*}

%% file: sections/appendix_protocol.tex
\section{Systems, Protocol, Cost and Latency}
\label{app:systems}

\paragraph{Systems.}
Table~\ref{tab:systems} lists the seventeen fast systems and the reasoning reference, grouped by type. Fourteen
are System One models, which return a native distribution over the supplied options in one pass. Two are hosted
APIs: JEV and OpenAI Decisions \citep{openai2026decisions}. Ten are released open-weight models that we serve
locally, pinned to released revisions, each with its authors' own serving code and documented settings:
Clef-27B, Kev-9B, CLM-8B, Laya, Eikos-27B, Perplexity Decider-27B, SimpleJev-27B, Vega-27B, Torchcast-27B and Quyet-1.0-Large.
Two are published recipes that read a frozen open-weight language model through the same interface: Cygnet
reads masked single-letter probabilities of Gemma-4-12B, scores long option lists in groups of 20 and calibrates
the result, and Decider-chat-31B reads the answer-letter distribution of Gemma-4-31B-it with a temperature that
depends on the number of options. The three open-weight language models are scored by complete-option likelihood
with thinking disabled. We render the state, question and enumerated options with the model's chat template, sum
the log probabilities of each option's full name followed by the end-of-turn token, and normalize over the
options. This exposes a full distribution that is directly comparable in form to those of the System One models.
A hosted reasoning model (GPT-6.1 Sol, medium reasoning) answers each question in a separate call with
the selected option and a verbalized distribution. It serves as a reference with deliberation, not as a
competitor to the System One models; it answered all 174,513 questions of the benchmark (322 answers could not be
parsed and count as wrong). Further released System One models are being evaluated with the same client,
request format and run records; we report only systems for which every record and every condition is complete.

\paragraph{Identical inputs.}
Every system receives the same frozen request: the same state, question text,
options, descriptions and option order. We never truncate a state, drop
options or reselect records to fit a system. When a request exceeds a
system's documented input limit, the question is recorded as unsupported,
counted wrong, and reported as coverage. This affects Laya (512 input tokens,
and option lists longer than its 192-token head budget), Cygnet (726 records
over its 16k context) and Gemma (4 records over 32k). It also affects
Quyet-1.0-Large, whose runtime refuses questions with more than ten
options, which includes every ALFWorld decision: only 3.8\% of paired test states
have all of its answers supported. Perplexity Decider-27B refuses questions whose
prompt exceeds 8,192 tokens (9.3\% of paired states have a refused answer, mostly
in $\tau^2$-bench and the visible-field probes); Decider-chat-31B and OpenAI Decisions
refuse 0.3--0.4\%. Vega-27B's server runs out of memory on the longest states on one 80\,GB H100;
after three failures on a healthy server we record such a question as unsupported (9,237 answers; 5.0\% of
paired states), and SimpleJev-27B's server fails repeatedly on the longest states (13,119 answers recorded
as unsupported after the same rules; 6.6\% of paired states). Two author runtimes shorten long states themselves before
reading them (Table~\ref{tab:systems}); we report them as shipped. Readouts differ across
systems by design, so comparisons are between deployed choice interfaces.
They do not isolate the effect of architecture, model size or training data.
Where an analysis uses a subset of questions, a system that answers fewer than half of them is left out of
that analysis and named (the \emph{coverage rule}).

\paragraph{What is evaluated.}
All seventeen fast systems and the reasoning reference receive all 65,228 records, including every paired test
state. The trace-ablation condition was run for all seventeen fast systems on the 10,897 original agent
records that carry it. Closed-loop
execution, in which a System One model acts as the policy in a live
environment, is outside the scope of this paper's results.

\paragraph{Latency.}
\label{app:latency}
Accuracy runs batch requests for throughput, so their timings are not latency.
We therefore time a separate fixed \emph{latency panel} of 163 test items, one per main
difficulty cell, so that every group, context length (37 to 99k characters),
question count (1--8) and option count (2--255) is represented. Each system
answers all questions of an item in one request, strictly sequentially, after
five untimed warm-up items. We run three passes in reshuffled order and report
the first pass, excluding the warm-up items; repeated passes, which benefit from
caches, are reported separately. Only complete answers count toward latency,
and refusals are reported as coverage. For pairs of systems we also report the
median of per-request time ratios on the requests both answer, which is more
robust than a ratio of medians. Seven systems, marked in Table~\ref{tab:systems}, were not timed on this panel. Their
accuracy runs, unlike those of the other systems, send one request at a time to a server on its own H100 (or,
for OpenAI Decisions, to the hosted API), so we estimate their latency from those runs on the same 158 timed
panel items; these estimates lack the panel's warm-up and are marked separately in Figure~\ref{fig:pareto}. The reasoning
reference answers one question per request, so we time it through its synchronous API on the same panel and order,
sending the questions of an item in parallel; all 158 timed items were answered, with a median of 2.3\,s per item,
about eleven times JEV's.

\paragraph{Cost.}\label{app:efficiency}
System One models emphasize intelligence per dollar, so we report cost
next to accuracy (Figure~\ref{fig:pareto}). The hosted JEV model was billed
for 209.2M input tokens over all 65,228 records, \$8.79 in total or \$0.13 per
1,000 decision states. For self-hosted systems we estimate cost from the panel's time
per state at a stated GPU price; all were timed on the same hardware class,
one request at a time (for the self-hosted systems not timed on it, from their
accuracy runs; \S\ref{app:latency}). On these estimates the efficient frontier
runs from Laya and Kev-9B through Cygnet, Decider-chat-31B and JEV to Eikos-27B
and Gemma-4-31B. Eikos-27B reaches 0.795 at an estimated \$0.19 per 1,000
states, about 1.4 times JEV's measured cost, and displaces Clef-27B from the
frontier; moving from JEV to Gemma raises paired decision accuracy from 0.763 to
0.811 at about five times the cost per state, while the Qwen readouts and CLM
are dominated. The second hosted decision model, OpenAI Decisions, was billed
for 186.9M input tokens on the 65,228 records (\$18.69 at \$0.10 per million
input tokens, output free), \$0.29 per 1,000 decision states or about twice
JEV's cost, for a paired decision accuracy of 0.706. The reasoning reference is shown for orientation only: its full run cost \$642.00 under batch
pricing, \$9.84 per 1,000 decision states or about 73 times JEV, for a paired decision accuracy of 0.884
(\S\ref{sec:dissociation}).

\paragraph{Model versions and reproducibility.}
Open models are pinned to released revisions and served with recorded
configurations. JEV is pinned by its version string
(\texttt{jev-1.13.0}), but its provider does not promise long-term support for
a version, and hosted serving can change or fail: during one check on
5~October~2026, 16 of 24 minimal requests returned server errors. We
therefore re-ran the decision question alone on a stratified sample (up to 25 per cell) of 2,783
paired test states on 6~October, after the formal runs had finished on
1~October. The selected action matched the formal run on 96.3\% of states
(2,681/2,783). Because the formal requests also carried the probe questions,
this agreement bounds the combined effect of version drift, serving
nondeterminism and request packing; it does not isolate any one of them.
OpenAI Decisions is a public beta identified only by its model name
(\texttt{gpt-6-luna}); all of its records were collected on 11~October~2026. We keep all
raw requests, full output distributions and timestamps, so that
later versions can be compared on identical inputs.

%% file: sections/appendix.tex
\section{Label audit: co-located receptacles}
\label{app:audit}

\input{tables/label_audit}

ALFWorld can place several receptacles at one native location (for example a countertop and a coffee
machine). The builder of the original cohorts mapped the agent's location to one representative receptacle,
chosen by fact order, and derived the location probe, the sightings and the set of visited receptacles from
that mapping. We re-derived all three from the text of the trace with ALFWorld's fixed message patterns
(``You arrive at R'', ``On the R, you see'', ``You open the R'', ``You pick up'', ``You move'', ``You
cool/heat/clean'', ``Nothing happens''), without simulator facts. Table~\ref{tab:audit} lists the
disagreements. In every inspected location case, the text states that the agent arrived at a different
receptacle from the label; for the 303 paired test records with such a label, JEV, Qwen3.8-27B, Cygnet and
Gemma answer the receptacle named in the text in 297--301 cases. The extra gold options are lenient rather
than wrong, since a co-located receptacle can be acted on from the same location. Among the observation minimal pairs, most seen-at
disagreements are edited sightings that a later visit to the same receptacle contradicts. The confirmation
cohort already uses set-valued location labels. We keep the frozen labels; excluding the 398 records
with a location or seen-at disagreement raises, for every strong system, the share of ALFWorld decision
errors made with all measured facts correct (Table~\ref{tab:auditsens}).

\begin{table}[h]
\centering
\small
\begin{tabular}{@{}lrrr@{}}
\toprule
System & All & Clean & Flagged \\
\midrule
Gemma-4-31B & 0.810 & 0.885 & 0.251 \\
Qwen3.8-27B & 0.792 & 0.864 & 0.295 \\
Cygnet & 0.776 & 0.834 & 0.334 \\
Clef-27B & 0.682 & 0.751 & 0.212 \\
JEV & 0.663 & 0.707 & 0.364 \\
\bottomrule
\end{tabular}
\caption{Share of ALFWorld decision errors made with all measured facts correct, on all 8,197 paired
ALFWorld test records, on the 7,799 records without a location or seen-at disagreement, and on the 398
flagged records.}
\label{tab:auditsens}
\end{table}

\section{Decision-oriented probes: definitions}
\label{app:dprobes}

All probe answers are derived from the task sentence, the initial observation and the trace. For each
state we ask: the agent's location (all receptacles plus ``the middle of the room''); what it holds (all
objects seen plus ``nothing''); for every object type the task still needs, where an instance was last seen
that is still available (all receptacles plus ``not seen yet''); whether the held target object has already
been cleaned, heated or cooled; how many target objects are already in place (two-object tasks); for every
receptacle, whether the agent has gone there; and the next sub-goal. Sub-goal categories are defined by
mapping each admissible action through the parsed state: search (an unvisited receptacle), fetch (a
receptacle where a needed object was seen), open the receptacle here, take a needed object, go to the
appliance, process, go to the destination, place, use the lamp, and two residual categories (another object;
look, examine or inventory). The gold sub-goal is the set of categories of the gold actions. States whose
gold actions fall into a residual category or into a revisit of a searched receptacle were excluded; all
inspected cases were co-located receptacles (Appendix~\ref{app:audit}). Of 705 sampled ALFWorld states,
644 remain. The probe request contains 26 questions per state on average.

\section{Generated procedures}
\label{app:ladder}

The generator produces states in a household domain (a home robot following a house protocol) and an
orders domain (a support agent following a refund policy). Each procedure is a decision list over four rule facts with a default action; the trace establishes eight
binary facts (the four rule facts and four distractors) through observations, some of which are later
overwritten. The action set is fixed per domain and includes actions licensed by
the distractor facts, so a model that acts on any true fact can be wrong. Sufficiency (a unique gold action
for every assignment of the four rule facts) is verified by enumeration. Dev seeds were used only for template checks; the 1,200 test states use disjoint seeds.

%% file: tables/label_audit.tex
\begin{table}[t]
\centering
\small
\begin{tabular}{@{}lrrrr@{}}
\toprule
 & & \multicolumn{3}{c}{Label disagrees with the text} \\
\cmidrule(l){3-5}
Group & Records & Location & Seen-at & Extra gold \\
\midrule
A1 & 3,000 & 140 & 23 & 232 \\
A2 & 300 & 11 & 0 & 47 \\
A3 & 300 & 22 & 0 & 19 \\
A4 & 1,500 & 54 & 20 & 0 \\
A5 & 997 & 57 & 70 & 58 \\
A7 & 300 & 19 & 0 & 35 \\
C3-ALF & 1,800 & 0 & 0 & 13 \\
\bottomrule
\end{tabular}
\caption{ALFWorld test records whose labels disagree with the trace text because receptacles share a simulator location. Location: the probe names a co-located receptacle. Seen-at: the sighting differs. Extra gold: the gold also contains an already searched receptacle (lenient).}
\label{tab:audit}
\end{table}